\documentclass[12pt]{article}
\usepackage{scicite} 
\usepackage{times}
\usepackage{graphicx} 
\usepackage{xcolor} 
\usepackage{placeins} 
\usepackage{multirow} 
\usepackage{caption} 
\usepackage{siunitx}
\usepackage{enumitem} 
\usepackage{lipsum}
\graphicspath{{Figures/}} 

\newenvironment{sciabstract}{
\begin{quote} \bf}
{\end{quote}}

\newcounter{lastnote}

\title{\LARGE \bf Stretchable Arduinos embedded in soft robots 
}

\author{Stephanie J. Woodman$^{1}$, Dylan S. Shah$^{1}$, Melanie Landesberg$^{1}$,\\ Anjali Agrawala$^{1}$, and Rebecca Kramer-Bottiglio$^{1}*$
\\
\normalsize{$^{1}$Department of Mechanical Engineering \& Materials Science, Yale University.}
\\
\normalsize{9 Hillhouse Ave, New Haven, CT 06511, USA.}
\\
\normalsize{*To whom correspondence should be addressed; E-mail:  rebecca.kramer@yale.edu.}
}

\makeatletter
\newcommand*\titleheader[1]{\gdef\@titleheader{#1}}
\AtBeginDocument{%
  \let\st@red@title\@title
  \def\@title{%
    \bgroup\normalfont\large\centering\@titleheader\par\egroup
    \vskip1.5em\st@red@title}
}
\makeatother

\titleheader{Notice: ``This is the author's version of the work. It is posted here by permission of the AAAS for personal use, not for redistribution. The definitive version was published in Science Robotics (Vol. 9 Issue 94, Sept. 2024), DOI: 10.1126/scirobotics.adn6844. ISSN (electronic): 2470-9476''.
\\
URL: www.science.org/doi/10.1126/scirobotics.adn6844}

\date{}
\begin{document}

\maketitle

\begin{sciabstract}
To achieve real-world functionality, robots must have the ability to carry out decision-making computations. However, soft robots stretch and therefore need a solution other than rigid computers.
Examples of embedding computing capacity into soft robots currently include appending rigid printed circuit boards (PCBs) to the robot, integrating soft logic gates, and exploiting material responses for material-embedded computation. 
Although promising, these approaches introduce limitations such as rigidity, tethers, or low logic gate density. 
The field of stretchable electronics has sought to solve these challenges, but a complete pipeline for direct integration of single-board computers, microcontrollers, and other complex circuitry into soft robots has remained elusive.
We present a generalized method to translate any complex two-layer circuit into a soft, stretchable form. This enabled the creation of stretchable single-board microcontrollers (including Arduinos) and other commercial circuits (including Sparkfun circuits), without design simplifications. As demonstrations of the method's utility, we embed highly stretchable ($>$300\% strain) Arduino Pro Minis into the bodies of multiple soft robots. This makes use of otherwise inert structural material, fulfilling the promise of the stretchable electronics field to integrate state-of-the-art computational power into robust, stretchable systems during active use. 
\end{sciabstract}

\section*{One Sentence Summary}%
Stretchable open-source electronics, like Arduinos, enable embedded computation in soft robots and garments. 
%

\section*{INTRODUCTION}
Arduinos and other single-board microcrontrollers are reprogrammable, widely used, and have an open-source community comprising over 1 million users~\cite{noauthor_arduino_2021}, making their prevalence as controllers unparalleled in and out of the robotics world. Most soft robots today are controlled by Arduino-style microcontrollers~\cite{noauthor_arduino_nodate,tolley_resilient_2014,Nemitz2016,Marchese2014,Stokes2014}. The modulus mismatch between rigid Arduino boards and the materials used in soft robots (for example, silicone elastomers) leads many designers to either place the electronics in regions of the robot designed to experience minimal strain~\cite{tolley_resilient_2014} or off-board them entirely~\cite{shah_morphing_2019, shepherd_multigait_2011, seok_meshworm:_2013}. Proposed solutions to this problem include mechanical computing platforms, soft logic gates, and stretchable electronics.

\begin{figure*}
     \centering
     \includegraphics[width=6in]{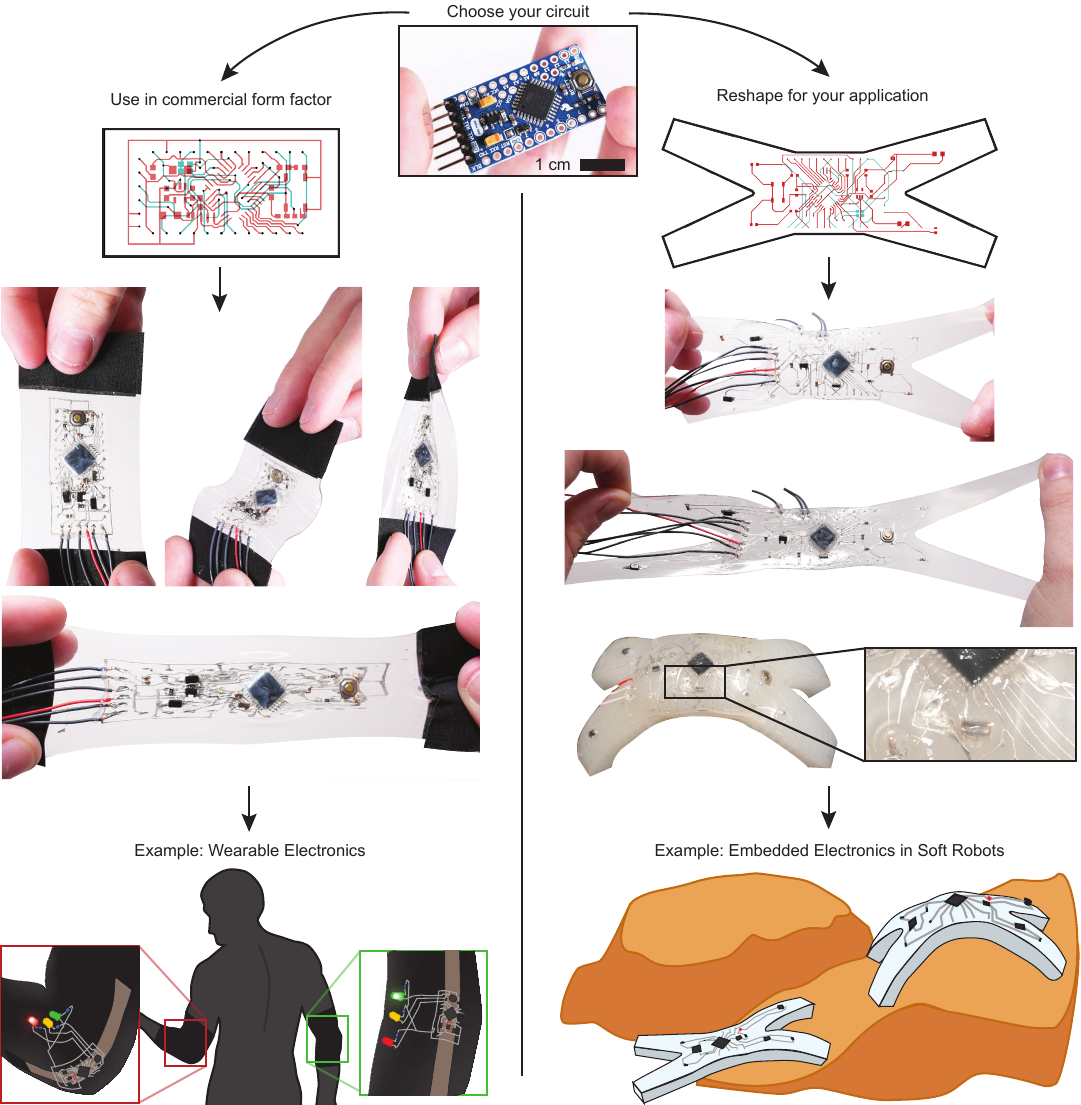}
     \caption{\textbf{End-to-end method to translate complex commercial circuit boards into stretchable forms.} The method is uniquely enabled by biphasic conductors and tackified silicones, enabling circuit integration into fully functioning wearables and soft robots during use.     
     }
     \label{fig: summary}
\end{figure*}

Instead of relying on conventional electronics, researchers have proposed fully soft mechanical computing platforms that exploit high-dimensional dynamic phenomena to perform computation and act as distributed information processing networks~\cite{yasuda2021mechanical}. Most of these non-von Neumann computational architectures take the form of tunable mechanical metamaterials that switch between two discrete states, analogous to binary digits (bits) in electronic computers. Soft logic gates have been proposed based on fluidic principles (pneumatic~\cite{drotman_electronics-free_2021, preston_digital_2019}, liquid~\cite{wehner_integrated_2016,Wissman2017}), pointing toward fully autonomous open-loop behaviors such as locomotion and arm motion. However, these mechanical approaches lack the logic gate density to match the capabilities of traditional electronic computers, which have proliferated due to miniaturization and scaling.

Another approach toward embedded computation in soft robots is the field of stretchable electronics, where researchers aim to endow traditional computing platforms with stretchability. This approach retains the computational power density of traditional circuitry while introducing stretchable conductive traces and substrates that link silicon-based rigid integrated circuits (ICs). 
Stretchable traces made from geometric patterning of solid thin-film conductors (for example, serpentines, meshes) are known for excellent interfacing with rigid microelectronics~\cite{wu_stretchable_2019, kim_stretchable_2008,Huang_3D_circuit_2018,Biswas_geometric_2019}, but typically cannot reach the strains required for soft robot applications (20-1000\% strain ~\cite{Rich2018}). 
In contrast, stretchable traces made from liquid metals (LM)~\cite{kramer_wearable_2011, dickey_stretchable_2017}, liquid metal composites~\cite{kazem_soft_2017, bartlett_high_2017, kazem_extreme_2018}, or conductive elastomers~\cite{yu_materials_2017,Lee-omnidirectional-2023} can often reach higher strains, but, first, have difficulty interfacing with rigid IC components and, second, display strain-dependent electrical resistances. 

Toward the former, recent attempts have utilized acidic or alkaline solutions to improve the electrical conductivity of the IC-LM interface~\cite{ozutemiz_scalable_2022, ozutemiz_egainmetal_2018}. 
Tang \textit{et al.}~\cite{tang_skin_2022} presented an LM-adhesive mixture to promote substrate adhesion and IC interfacing, yet the inclusion of adhesive greatly reduced the formulation's conductance. Other works highlight how the interfaces between LM-based conductive inks and commercial ICs often suffer when strained, thus confining examples to low-strain demonstrations~\cite{Tavakoli2022,Lopes_silver_2021,tang_skin_2022,Lopes-santos-2021,Lee-universal-2022, ozutemiz_scalable_2022,Hellebrekers2018,Yin2020,Xie2023} or
simplified designs~\cite{ozutemiz_scalable_2022,ozutemiz_egainmetal_2018,Liu_Bgain_2021,Lee-universal-2022}. 
For example, Valentine \textit{et al.}~\cite{valentine_hybrid_2017} proposed a non-commercial-form single-layer microcontroller circuit, but characterized only a simpler LED circuit with LM interconnects under strain.

Toward the latter, biphasic (solid-liquid) gallium-based formulations have been proposed as ``strain-insensitive'' conductors. Most biphasic metal formulations are made from a mixture of liquid eutectic gallium-indium (EGaIn) and solid metal particles (such as silver~\cite{Tavakoli_silver_2018, Lopes_silver_2021, Wang_silver_2018}, copper~\cite{Li_copper_flakes_2020}, and nickel~\cite{daalkhaijav_rheological_2018}) and yield favorable rheological properties for patterning, interfacing, and enhanced electrical performance. Formulations with added or natively-formed semiconductive particles such as quartz~\cite{Chang_quartz_2020}, graphene oxide~\cite{lee_self-mixed_nodate}, and gallium oxide~\cite{Liu_Bgain_2021, Sanchez-Botero2022} boast a suppressed strain response---an especially promising attribute for stretchable circuitry.
We previously used biphasic gallium-indium (BGaIn), made from EGaIn with \textit{in situ}-formed crystalline gallium oxide (Ga$_2$O$_3$) growth ($\sim$34 wt\% solid), to achieve high conductivity (2.06$\times$10$^6$~S/m), extreme stretchability (up to 1,000\%), suppressed resistance changes when strained, cyclic stability (consistent performance over 1,500 cycles), and a reliable interface with rigid electronics~\cite{Liu_Bgain_2021}.
However, prior biphasic metal formulations were found to be laborious to produce at scale.

In this work, we present material and interfacial processing solutions to the prior challenges, which enable a generalized end-to-end method to translate any two-layer circuit (such as an Arduino) into a soft stretchable form ready for integration into soft robots and wearable electronics (Fig. 1 
and Movie 1).
This method is made possible by, first, a scalable stretchable conductor with a suppressed strain response, cyclic stability, and reliable interfacing to ICs; and, second, derived guidelines for conductor-substrate compatibility, which determines the reliability of the process.
To demonstrate utility, we translate several open-source circuit designs containing ICs with dozens of interfaces and vertical interconnect accesses (VIAs) into stretchable forms, and characterize their function during high-strain cycling.
We create a soft, stretchable Arduino Pro Mini that can be 
stretched to $>$300\% strain and maintains functionality over $>$120 cycles to 100\% strain, despite having more than 70 interfaces between rigid and soft components and more than 40 VIAs. 
We further demonstrate method generality by fabricating stretchable versions of the Arduino Lilypad, Sparkfun Sound Detector, and Sparkfun RGB (Red Green Blue) and Gesture Sensor. Notably, the circuits utilize the same form factor and IC packages as the originals, including both no-lead (for example, dual flat no lead, DFN) and leads (for example, thin quad flat pack, TQFP).
Finally, we embed stretchable Arduino Pro Minis into the bodies of soft robots at specifically high-strain 
locations, and use them for embedded computation.  
The demonstrations collectively mark a transition from one-off, functionally limited showcases to robust, reliable, and complex multilayer stretchable circuits.

\section*{RESULTS}
\subsection*{Scalable, stretchable conductor with suppressed strain response}
Recently, there has been much interest in so-called strain-insensitive stretchable conductors, which have a suppressed strain response relative to bulk conductor assumptions. In contrast, most traditional circuits employ annealed copper traces (resistivity 1.72$\times$10$^-8$~$\Omega$m, density 8.93 g/cc at 20$^{\circ}$C).
As a classical constant-conductivity bulk conductor, copper theoretically follows Pouillet’s law~\cite{zhu_ultrastretchable_2013} when strained: 
\begin{equation}
R/R_0 = (1+\epsilon)^{1+2\nu}
\end{equation}
where $R/R_0$ is the relative (normalized) resistance
change, $\epsilon$ is the applied strain, and $\nu$ is the Poisson’s ratio, which is most often assumed to be 0.5, implying the conductor is incompressible~\cite{Liu_Bgain_2021}.
Although Pouillet’s law predicts large increases in resistance even at moderate strains for bulk conductors, by engineering materials with stable conductance under large strains, stretchable complex circuits and low-loss power transmission can be achieved. 
Our prior work presented one such conductor with an electromechanical response supressed far below that predicted by Pouillet's law---a BGaIn composition formed from semiconductive crystalline gallium oxide particles mixed with EGaIn~\cite{Liu_Bgain_2021, Sanchez-Botero2022}. However, BGaIn is labor-intensive to produce and the synthesis process yields small quantities per batch.

Another proposed biphasic conductor is oxidized gallium-indium (OGaIn;~\cite{Kong_Ga_foam_2020, Wang_ogain_2019}), a biphasic foam containing amorphous gallium oxide particles and EGaIn. OGaIn is made via rigorous mixing of EGaIn in air, and can be created at scale. 
In this study, we sought to rigorously characterize OGaIn in an attempt to replicate the desirable properties of BGaIn with a material that can be manufactured at a scale useful to industry. Interestingly, despite the OGaIn and BGaIn materials having similar qualitative rheological characteristics, we measured OGaIn to be only 1.4 wt\% amorphous gallium oxide (close to the 1.21 wt\% reported by Chen \textit{et al.}~\cite{Chen2023}), whereas BGaIn is $\sim$34 wt\% solid crystalline gallium oxide~\cite{Liu_Bgain_2021}. We surmise that OGaIn is primarily thickened through air inclusions~\cite{Wang_ogain_2019}, which are not accounted for in wt\% formulations. Indeed, the density of OGaIn (4.65 g/cc at 20~$^{\circ}$C) was substantially lower than EGaIn's ($>$6 g/cc). Scanning electron microscope (SEM) micrographs, energy dispersive X-ray spectroscopy (EDS), and X-ray Diffraction (XRD) analysis (Fig.~S1
, Supplementary Methods) suggested that the hardness of the gallium oxide that forms at the OGaIn-air interface prevents deformation of the air inclusions upon stretching, which could explain the observed viscosity enhancement of OGaIn relative to neat EGaIn. We further investigated the rheological properties of OGaIn, revealing its shear-thinning behavior, which suggests compatibility with extrusion printing techniques (Fig.~ S2).
Finally, OGaIn and BGaIn exhibited similar bulk electrical conductivities of 2.11$\times$10$^6$~S/m and 2.06$\times$10$^6$~S/m~\cite{Liu_Bgain_2021}, respectively. 

\begin{figure*}
    \centering
    \includegraphics[width=136mm]{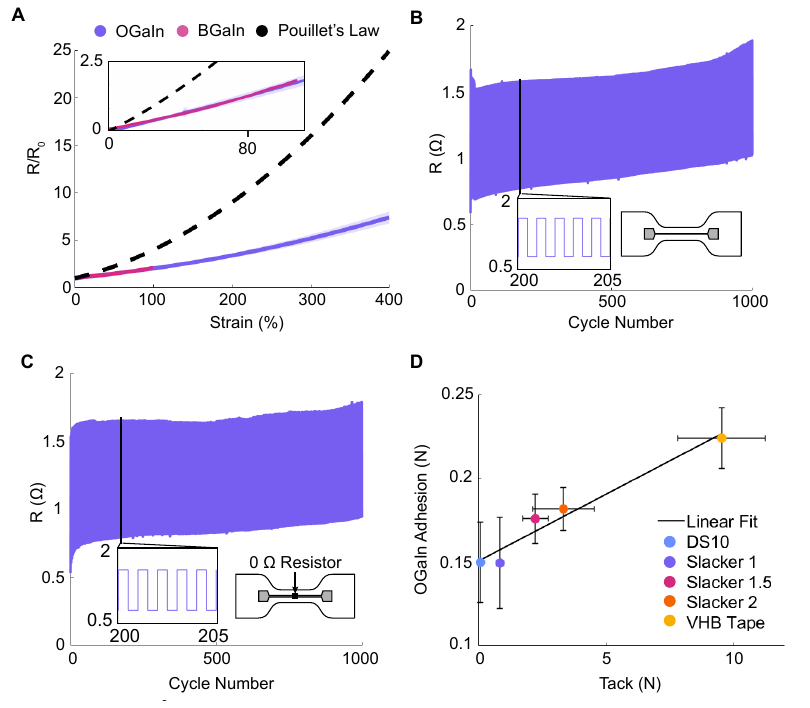}
    \caption{\textbf{OGaIn characterization and substrate compatibility.}
    (\textbf{A}) Normalized resistance change vs. strain for 5 OGaIn samples, plotted alongside the theoretical values for a bulk conductor and previous results from our biphasic gallium-indium alloy (BGaIn, ~\cite{Sanchez-Botero2022}). Shaded region represents one standard deviation.
    (\textbf{B}) Representative electrical resistance vs. cycle number for an OGaIn trace stretched to 150\% strain over 1000~cycles.
    (\textbf{C}) Representative electrical resistance vs. cycle number for an OGaIn trace with a 0~$\Omega$ resistor stretched to 150\% strain over 1000~cycles.
    (\textbf{D}) OGaIn adhesion vs. ASTM D6195-22 tack measurement of various stretchable substrates. Error bars indicate one standard deviation from the mean over 5~trials.
    }
    \label{fig: OGaIn char}
\end{figure*}

To assess the electromechanical performance, we patterned 250~$\mu$m traces of OGaIn (as wide as the thinnest trace in the commercially-available Arduino Pro Mini, Fig.~S3
A, Materials and Methods) onto an ASTM (American Society for Testing and Materials) D412~\cite{d11_committee_d412:_2016} standard dogbone shape of acrylic foam tape (VHB (Very High Bond) Tape, 3M) that adheres strongly to IC packaging (Fig.~S3
B,C), and encapsulated the traces with rubber cement (Elmers Inc.). 
We then strained each sample to 400\% strain (limited by our test setup). 
We found that the electromechanical response of OGaIn is below that predicted by bulk-conductor assumptions (Pouillet's Law, Fig~2
A)~\cite{Liu_Bgain_2021, zolfaghari_network_2020, zhu_ultrastretchable_2013}. As one point of comparison, OGaIn has an $R/R_0$ of 7 at 400\% strain, compared to 25 for a bulk conductor. Further, the $R/R_0$ vs $\epsilon$ behavior of OGaIn matches previously-reported values for BGaIn in the range where comparison was possible (up to 100\% strain)~\cite{Sanchez-Botero2022}. To provide a more direct comparison, we additionally tested OGaIn and BGaIn on the same substrate 
and with the same tracewidth 
(Fig.~S4
), finding similar agreement. OGaIn is thus a scalable replacement for BGaIn.

To evaluate cyclic stability, we subjected single-trace samples (unstretched resistance $\sim$0.5~$\Omega$, which is consistent with prior literature~\cite{lin_design_2017,marques_reliable_2019}) to 1000 cycles of 150\% strain at 15 mm/min (Fig.~2
B). 

After an initial break-in period of a few cycles~\cite{Bueche1960}, the trace resistance increased by only $\sim$0.5~$\Omega$ between cycles 5 and 1000. 
To evaluate interfacing stability between traces and rigid ICs' contact pads, we repeated the cyclic strain (Fig.~2
C; Fig.~S3
E) and high strain (Fig.~S5
) tests with two traces bridged by a 0~$\Omega$ resistor (Digi-Key, Inc.). There were no discernible differences between the interfaced and non-interfaced traces, whether comparing the initial raw resistance or the resistance after cycling. This, along with contact resistance measurements (Fig.~S6
), indicates that the slight increases in resistance observed with cycles are not attributable to interfacing. Possible sources of increased resistance may include an elongation of the trace length after stretching (due to viscoelastic properties and plastic deformation of the surrounding polymers) or a change in OGaIn's effective electrical conductivity (for example, through phase separation).

\subsection*{Deriving guidelines for conductor-substrate compatibility}
The wetting and adhesion of liquid metals and biphasic metals to a host substrate is an important parameter in stretchable electronics applications~\cite{joshipura2021contact,Cook_peck_2019}. In this work, we rely on the adhesion of OGaIn to both its underlying substrate and its co-located ICs. Given the cyclic stability shown in Figs.~2
B-C, we inferred that this adhesion is sufficient on the VHB tape. However, toward our goal of embedding complex stretchable circuits into the bodies of soft robots, we sought to characterize the adhesion of OGaIn to other materials commonly used in soft robotics, in addition to probing how to modify existing materials to improve adhesion. Further, we selected substrate materials to derive a generalized correlation between OGaIn-substrate adhesion and substrate material tack.

We compared a baseline VHB tape substrate to four silicone elastomer (DragonSkin 10, abbreviated as DS10; SmoothOn Inc.) substrates with increasing fractions of a tactile mutator (Slacker, SmoothOn Inc.). We refer to the neat silicone elastomer as DS10, the formulation with a mixing ratio of 1:1:1 (Part A : Slacker : Part B) as ``Slacker 1,'' and formulations with mixing ratios (1:1.5:1) and (1:2:1) as ``Slacker 1.5'' and ``Slacker 2,'' respectively. Further details can be found in Materials and Methods.

Our tests show a positive linear correlation (R$^2$ = 0.96) between substrate tack and OGaIn-substrate adhesion (Fig.~2
D).
All substrates had at least some OGaIn adhesion, even with $<$0.1~N tack, which is likely due to the oxide particle inclusions in OGaIn~\cite{joshipura2021contact} that encourage wetting~\cite{Cook_peck_2019}.
In practice, we find that substrates with low tack values (such as neat DS10 and Slacker 1 in Fig.~2
D) increase the likelihood of failure modes (such as trace defects and IC shifting, Table~S1
). Therefore, we recommend selecting substrates with a tack value of at least 0.18~N to ensure sufficient OGaIn-substrate adhesion, stable IC placement, and general conductor-substrate compatibility for stretchable circuits. %

\subsection*{Translating complex circuits to stretchable forms}
Having identified a suitable stretchable conductor and generalized its compatibility with soft, stretchable substrates, we then developed a method of translating as-is, complex circuit-board designs into stretchable circuits. Throughout the development of the method reported herein, we emphasized accessibility, aiming to eliminate a need for extensive equipment or materials expertise, or circuit design expertise. 
We applied the method to make stretchable versions of the popular Arduino Pro Mini, a reprogrammable single-board microcontroller, as well as the Arduino Lilypad, Sparkfun Sound Detector, and Sparkfun RGB and Gesture Sensor.

The circuits were fabricated using laser cutting and stencil printing (Fig.~3). 
The substrate (for example, VHB tape) was sandwiched between two layers of 0.1~mm thick sticker paper that act as masks. The main tradeoff we noted in selecting the sticker paper was that thinner paper allowed for more precise traces, whereas thicker paper facilitated easier removal and reduced smearing of traces due to ripping. %
The board outline and VIAs were cut using a carbon dioxide (CO$_2$) laser (Universal Laser VSL 2.30DT), and the trace outlines were etched on the bottom mask using a UV laser (LPKF Protolaser U4). OGaIn was painted onto the bottom traces, the mask was removed, and the circuit was encapsulated with a thin layer of rubber cement (Elmers, Inc.). Next, the top traces were made using the same laser and painting procedure, followed by component placement and a rubber cement sealing step. Finally, silicone adhesive (Sil-Poxy, SmoothOn Inc.) was added in a small region around the microprocessor, to reduce stress caused by the stiffness gradient. For full process details, we refer the reader to Materials and Methods. In contrast to our more restrictive prior art that utilized transfer printing~\cite{Liu_Bgain_2021}, this scalable, screen-printing and laser-cutting-based process enables the creation of encapsulated multilayer circuits using substrates that meet the compatibility requirement (tack $\ge\sim$0.18~N), and can accommodate dense IC components while withstanding high strains. Furthermore, increased adhesion led to increased robustness of the IC-OGaIn interfaces, allowing us to directly transfer commercial circuits into stretchable forms.
\begin{figure*}
    \centering
    \includegraphics[width=6 in]{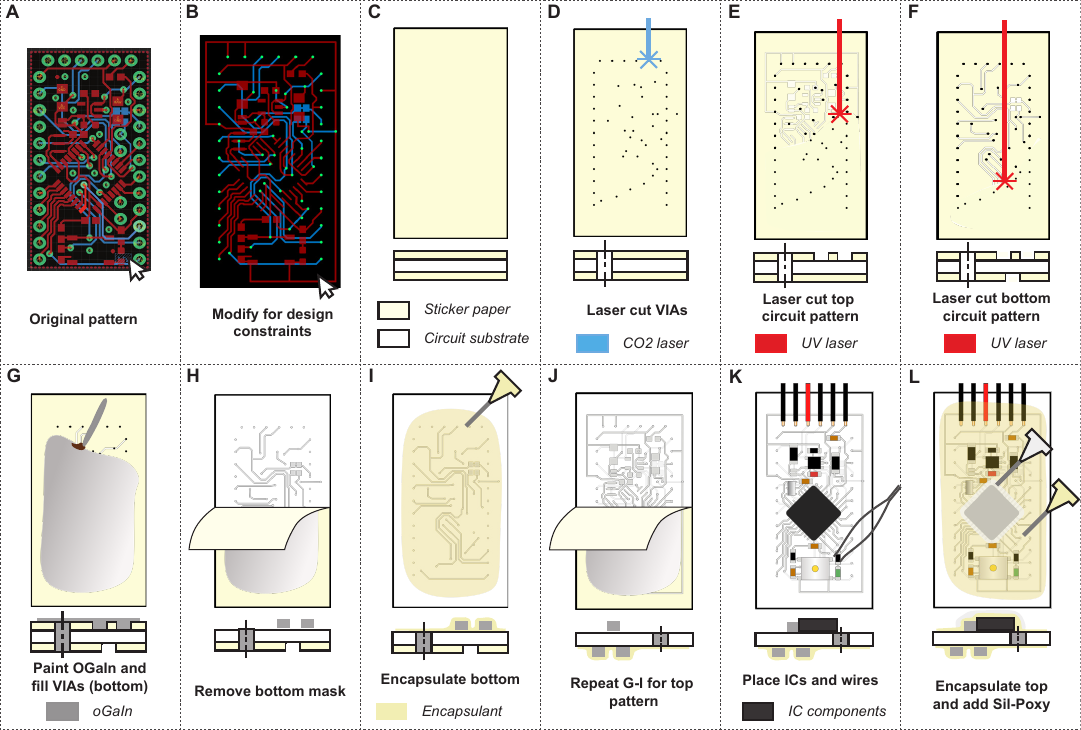}
    \caption{\textbf{Circuit manufacturing.} 
    (\textbf{A}) Original, open-source Arduino Pro Mini file in Autodesk Eagle.
    (\textbf{B}) Modified circuit design ready for laser cutting. 
    (\textbf{C}) Circuit substrate sandwiched between sticker paper.
    (\textbf{D}) Laser cut board outline and VIAs (vertical interconnect accesses) in sticker paper with CO$_2$ laser.
    (\textbf{E}) Cut top layer of circuit traces using UV laser. 
    (\textbf{F}) Cut bottom layer of circuit traces using UV laser. 
    (\textbf{G}) Paint bottom traces and fill VIAs with OGaIn.
    (\textbf{H}) Remove bottom mask and test trace conductivity. 
    (\textbf{I}) Encapsulate with material of choice. 
    (\textbf{J}) Paint top traces with OGaIn and remove mask. Test trace conductivity. 
    (\textbf{K}) Place components and wires. 
    (\textbf{L}) Encapsulate top. Add Sil-Poxy around microprocessor. 
    }
    \label{fig: mfg}
\end{figure*}
Straining the soft Arduino Pro Minis to failure (defined as a disconnection of serial communication to the computer, Fig.~4
A-C; Movie S1; Materials and Methods, strain rate 15~mm/min), we found that serial disconnect always occurred due to loss of electrical contact or shorting of the traces prior to mechanical failure of the substrate. The average strain at serial disconnect was 328\%, with the five samples failing between the range of 202\% strain and 404\% strain, corresponding to a failure in the plastic regime of each force-displacement curve (Fig.~4
D).
For two out of the five samples tested, the circuit no longer functioned after returning to the unstrained state. However, two samples re-booted in an error state (no serial communication), and one sample remained reprogrammable after being strained past initial serial disconnect. 

Cycling the Arduinos to 100\% strain (15~mm/min rate) demonstrated stability over at least 100 cycles, with the average number of cycles at failure being 124 and one sample reaching 200 cycles before failure (Fig.~4
E). For cyclic testing, circuits would be functional until an initial failure, and not recover after that first failure even when unstrained. 
Force vs. strain curves for neat substrates (VHB) and circuit-embedded substrates reveal that the circuits have a slight bulk stiffening effect on their host materials, although this appears to diminish with further cycling (Fig.~4
F). After 200 cycles, the materials exhibited similar behavior, with the force at 100\% strain only 27\% higher for the circuit compared with neat VHB. Both cycles exhibited similar plastic deformation after the first cycle (the Mullins effect~\cite{Mullins1969}).

\begin{figure*}
    \centering
    \includegraphics[width=6 in]{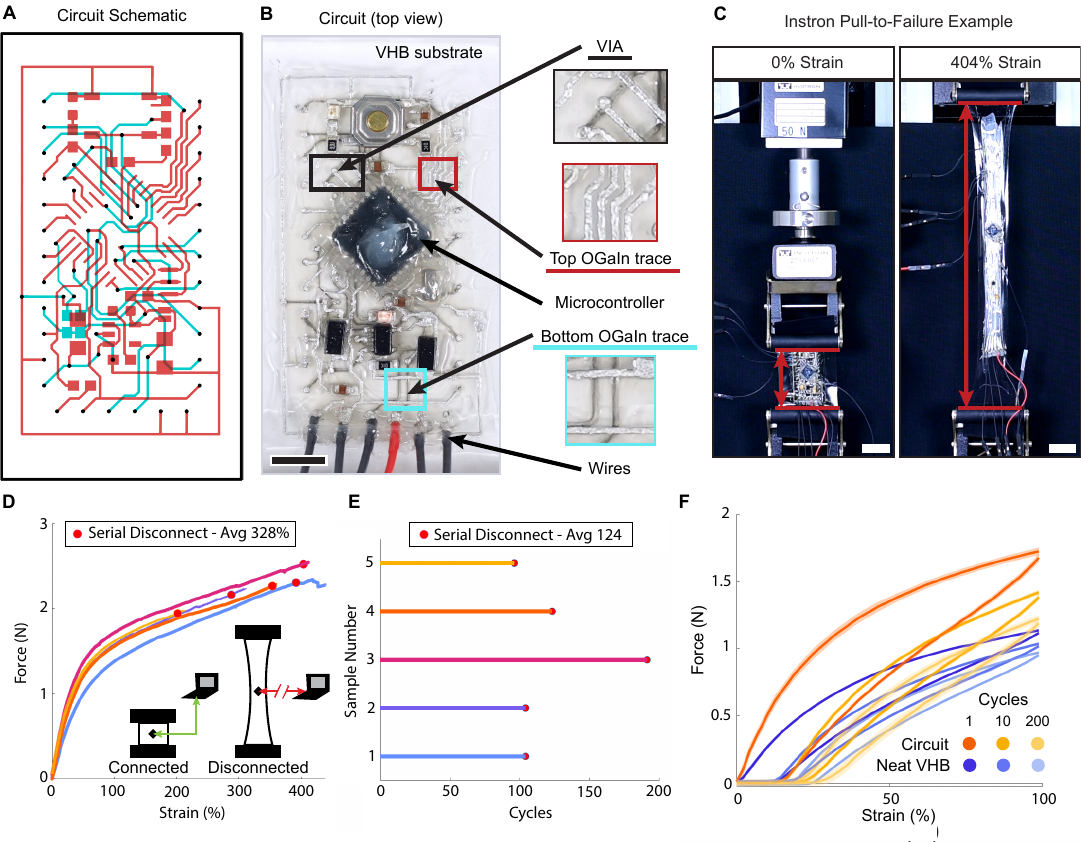}
    \caption{\textbf{Circuit characterization.} All circuits in this figure used VHB tape as a substrate.
    (\textbf{A}) Circuit schematic showing top and bottom traces in red and cyan, respectively, and vertical interconnect accesses (VIAs) in black.
    (\textbf{B}) Identifying key circuit components and materials. Note that the top and bottom interconnect layers can be seen, since VHB tape is translucent. In addition, the VIAs can be seen as round dots at various locations on the interconnects. Scale bar 5 mm. Trace widths in insets are 0.25 mm.
    (\textbf{C}) Image of circuit before strain testing on the materials testing system (Instron 3345) and just before serial disconnect at 404\% strain. Scale bars 18.8 mm.
    (\textbf{D}) Force vs. strain curves for each sample when strained until serial disconnect, noting when serial disconnect occurred for each sample. 
    (\textbf{E}) Cycle number when serial disconnect occurred for each sample, when the circuit was repeatedly strained to 100\%.
    (\textbf{F}) Comparing cyclic behavior of neat VHB tape (circuit substrate material) and the circuit at 1, 10 and 200 cycles. Solid lines are means over 5 samples after the number of strain cycles indicated in the legend. Shaded area indicates one standard deviation.
    }
    \label{fig: full circuit char}
\end{figure*}

Post-test analysis revealed that the strain-limiting Sil-Poxy region around the microprocessor, though it prevented shorting between microprocessor pins, introduced the most common failure mode of trace bridging at the edge of the strain-limiting region (Fig.~S7
). This result suggests that the soft circuits could be improved by introducing stiffness gradients into the substrate~\cite{bartlett20153d}. Another failure mode observed was at the trace-IC interface, indicating that enhancing the interfacial electrical connections could extend the operational strain range of the circuits. Additionally, two trace failures were observed during cyclic testing, seemingly unrelated to a rigid-soft interface. These failures are suspected to be caused by trace thinning due to viscoelastic effects.

To ensure generality with respect to commercial circuit design and IC package types, we stretched to failure one sample each of the Arduino Lilypad, Sparkfun Sound Detector, and Sparkfun RGB and Gesture Sensor (Fig.~5
, Movie S2). The Lilypad, whose rigid counterpart was specifically intended for wearable applications, strained to 415\% before slipping out of its mount while still functioning (Fig.~5
A). The Sound Detector, which is an analog circuit containing a through-hole mounted microphone, strained to a lower strain of 258\% (Fig.~5
B). Finally, the RGB and Gesture Sensor only uses no-lead IC packages, endowing it with higher stretchability, and it strained to 442\% without failure (Fig.~5
C).

\subsection*{Soft robots with embedded soft computers}
The soft Arduino Pro Minis were embedded into three soft robotic systems where: first, an Arduino controlled the gait of a locomotive robot; second, an Arduino sensed and communicated physical contact between agents in a multi-robot system; and third, an Arduino categorized the stretch data (low, medium, high) and visually indicated the user's motion in a wearable system (Fig.~6
). 

\begin{figure*} 
    \centering
    \includegraphics[width=68mm]{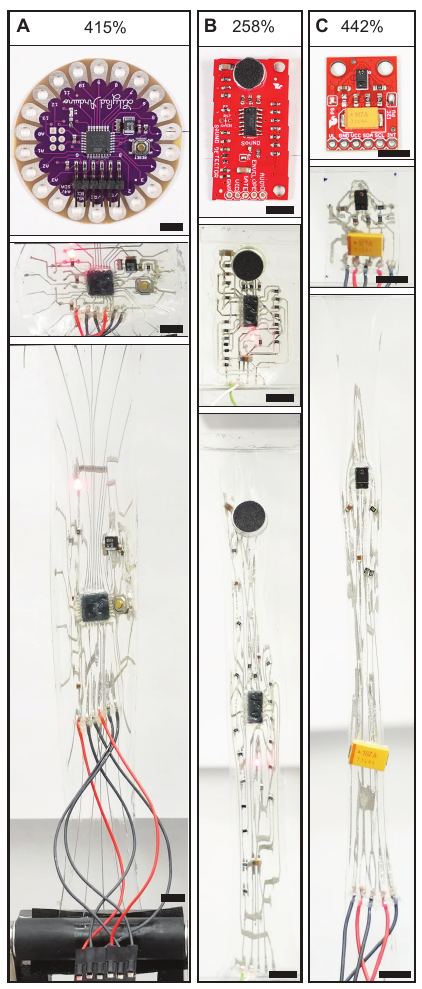}
    \caption{\textbf{Additional circuits.} 
    (\textbf{A}) Arduino Lilypad in rigid and soft form at 0\% strain, and the soft form at 415\% strain. Scale bar 7~mm.
    (\textbf{B}) Sparkfun Sound Detector in rigid and soft form at 0\% strain, and the soft form at 258\% strain. Scale bar 9~mm.
    (\textbf{C}) Sparkfun RGB and Gesture Sensor in rigid and soft form at 0\% strain, and the soft form at 442\% strain. Scale bar 9~mm.
    }
    \label{fig: other circuits}
\end{figure*}

\subsubsection*{\textit{On-board pneumatic control of a quadruped soft robot}}
We first aimed to demonstrate how soft single-board microcontrollers, and by extension other complex circuits, can be embedded into the structural material or unused surface area of soft robots---even at high-strain locations. We fabricated stretchable Arduino circuits on, first, a silicone substrate (Slacker 1.5) (Fig.~6
A), which matches the material of a silicone robot, and second, a VHB tape substrate (Fig.~6
B), and integrated them into a canonical quadrupedal soft robot form factor~\cite{shepherd_multigait_2011, tolley_resilient_2014} (Materials and Methods). In both cases, the microprocessor IC was integrated into the robot at the location of the highest strain during robot locomotion. The onboard soft Arduino was powered by a USB cable and used I$^2$C (inter-integrated circuit) to command off-board pressure regulators~\cite{booth_addressable_2018} to inflate the actuators and generate a forward walking gait (Fig.~6
C). The soft Arduino achieved successful processing and external communication at maximum strains of $\sim$100\% (Movies S3, S4). The soft Arduinos feature a custom layout in this demonstration, while keeping the most delicate components in the area of highest strain, to showcase that design modifications can be made to tailor-fit the host robot (Fig.~1
).

\begin{figure*}
    \centering
    \includegraphics[width=5.8 in]{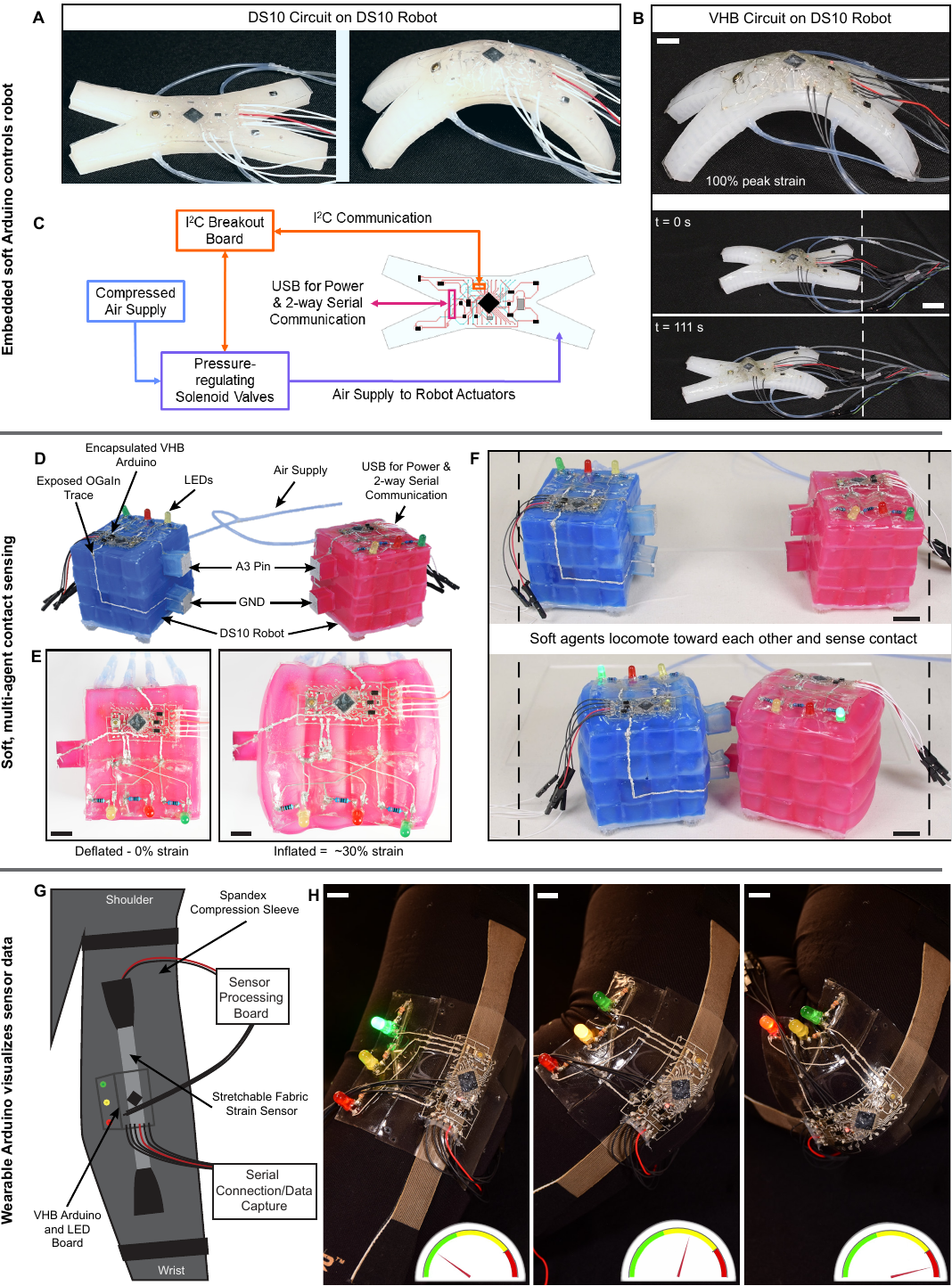} 
    \caption{\textbf{Embedded computation in soft systems.} Caption continued on next page. }
    \label{fig: demos}
\end{figure*}

\begin{figure}[t]
  Caption continued from previous page. (\textbf{A}) Silicone circuit embedded in a silicone (Dragon Skin 10, DS10) robot before it takes a step and mid-step, in a location where the circuit experiences $\sim$100\% strain.
    (\textbf{B}) A circuit with VHB as the substrate is mated to a DS10 robot at full inflation and at different times during the walking gait. Scale bar 8~mm.
    (\textbf{C}) Schematic of the quadruped driven by its embedded Arduino. 
    (\textbf{D}) Stretchable Arduinos used for contact sensing in a two-robot system.
    (\textbf{E}) Top view of the robot before and after inflation (before taking a step, and mid-step). Scale bar 8~mm.
    (\textbf{F}) Robots' initial poses and after they make contact. Scale bars 10~mm.
    (\textbf{G}) Stretchable Arduinos integrated with sensing circuitry in a smart garment that senses the user's elbow bending motions.
    (\textbf{H}) Sequential images showing a user bending their elbow and the Arduino detecting the motion as displayed by the gauge. Scale bars 7~mm.
\end{figure}

\subsubsection*{\textit{Neighbor-neighbor sensing in a multi-agent system}}
To illustrate scenarios where embedded computation could enable new capabilities in mobile soft robots, we integrated soft Arduinos into the surface of fully stretchable, pneumatic VoxelBots~\cite{kriegman_scalable_2020, kriegman_automated_2019}. Notably, the VoxelBots contain no strain-limiting surface, precluding rigid on-board PCB's and requiring all included circuitry to be stretchable. This multi-agent demonstration is a step toward our vision of a cell-like, modular approach to distributed computation, wherein robot modules may reconfigure relative to one another as a form of global shape change~\cite{shah_shape_2020}. 

The VoxelBots were created by casting hollow cubes of silicone (DS10), bonding them together using silicone, inserting pneumatic tubing, and attaching stretchable VHB-based Arduinos using DS10 (Materials and Methods). To implement a visual output, we additionally added a stretchable LED board (Fig.~S8
) connected to the Arduino (Fig.~6
D) embedded into the surface of the VoxelBot. Finally, to allow each VoxelBot to sense contact with other VoxelBots, one of the Arduino's analog pins was connected to an exposed OGaIn trace on one of two front faces on the robot. The other face, on the bottom, contained an exposed OGaIn trace connected to the Arduino's ground (GND) pin. 
The completed VoxelBots were then sequentially inflated and deflated using compressed air to locomote toward each other over approximately 30 steps (Fig.~6
E), as the onboard soft Arduino was programmed to sense and report contact with the other VoxelBot. When the input pin (A3) and GND pins on each robot made contact, their embedded Arduinos detected the voltage change and lit the VoxelBots' green LED to indicate successful contact (Fig.~6
F, Movie~S5).

\subsubsection*{\textit{Co-located computation and sensing in a wearable system}}
Given that the maximum strain of the stretchable Arduinos (on average 300\%) far exceeds the maximum strain experienced by the human body ($\sim$75\% strain~\cite{Chortos_skin_2016}), they can be placed virtually anywhere on smart garments. In contrast to prior works that use stretchable conductors as strain sensors and locate the IC interfaces in low- or no-strain locations~\cite{valentine_hybrid_2017, bartlett_rapid_2016, tang_skin_2022,Xie2023}, we intentionally placed our soft Arduino on the high-strain elbow skin (Fig.~6
G). We pressed a VHB-based Arduino and accompanying LED board (Fig.~S8
) onto a Spandex elbow sleeve (using the VHB tape and rubber cement's natural adhesion) with an embedded textile capacitive strain sensor~\cite{Anjali_wearable} (Fig.~6
G). Red, yellow, and green LEDs were attached to three pins of the stretchable Arduino (Materials and Methods).

The Arduino was programmed to convert voltages from the sensing board to capacitance and print capacitance values to the serial port. Simultaneously, the Arduino indicated the strain range by lighting up either the green, yellow, or red LEDs (Fig.~6
H, Movie~S6). The red LED indicated a high strain ($>$80 pF), yellow indicated a middle range (78-80 pF), and green indicated little or no strain ($<$78 pF). Capacitance increases with strain due to an increase in surface area of the parallel plate electrodes~\cite{Anjali_wearable}. Overall, this demonstration highlights the design freedom enabled by high-strain interfaces with respect to circuit placement.

\section*{DISCUSSION}
We presented a generalized, scalable, accessible, end-to-end method to translate complex two-layer circuits into soft, stretchable forms. We employed and characterized a biphasic stretchable conductor with a suppressed strain response, cyclic stability, and robust interfacing to ICs, and determined that substrates with tack $\ge$0.18~N are necessary for conductor-substrate compatibility. These revelations together allowed us to introduce a commercial-form stretchable Arduino.
We further made stretchable forms of other commercial circuits, from the popular vendors Arduino and Sparkfun, which show the breadth and generality of the method to a wide range of open-source, commercially-available designs. 

We integrated stretchable Arduino Pro Minis into several soft robotic systems in locations where they experienced $>$100\% strain. 
These integrations demonstrated soft locomotion quadrupeds with embedded electronics to control gait, a
fully soft two-agent robotic system using on-board, stretchable control and sensing,
and detection of a user's arm motion in a smart garment with the stretchable circuit intentionally placed in the area of the highest strain. 
The simplicity, accessibility, and success of the stretchable circuit fabrication method described herein give it the potential to be widely adopted by researchers, hobbyists, and industry. 
All of our circuits are based upon open-sourced designs, and our design files used to create the masks have been made available 
on our accompanying GitHub repository.

Our long-term goal is to develop stretchable versions of a wide range of familiar circuits (for example, within the Adafruit and Sparkfun communities), advancing the field of soft robotics and wearables and lowering the barrier to entry into soft robot research by enabling the use of common control platforms. Important functionalities could be added to make the robots more self-contained, such as Bluetooth modules and flexible/stretchable batteries to reduce external wiring requirements, and soft valves to bring additional pneumatic capabilities onboard. The process could be further tested to determine Gerber-like style guidelines, which could then be added into common design software (such as Eagle or Altium) to determine design viability in stretchable form. Additionally, we hope that this technique will be adopted by vendors of proprietary circuits, alongside an industry-academia joint effort to develop standardized connectors analogous to Qwiic (Sparkfun, Inc.) or USB connections from rigid electronics. 

Future work will include further improving maximum circuit strains, possibly through increased substrate tack or surface-treating the ICs with adhesion promoters. We also aim to reduce the manufacture time by using commercial stencil-printing machines for the traces, nozzle dispensing for VIAs~\cite{daalkhaijav_rheological_2018}, and automated pick-and-place machines for the ICs.
We hope this work will enable further research into soft systems endowed with computational intelligence rivaling today's rigid systems without sacrificing compliance, thus facilitating the realization of soft robots that can sense, decide, act, and adapt in the real world.

\section*{MATERIALS AND METHODS}
\label{sec: Methods}

\subsection*{OGaIn fabrication and characterization}
\label{methods: ogain fab char}
OGaIn was fabricated by measuring 200 mL of EGaIn into a 250 mL glass beaker and mixing using an IKA Eurostar 20 digital mixer, with a 4-blade stainless steel propeller, for 30 minutes (Fig.~S3
A). 
Resistivity and conductivity were estimated by measuring the resistance of a trace with known length, width, and height using a 4-point probe multimeter (BK Precision).
The weight percent of solid particles was measured by weighing a sample of OGaIn before and after dissolving the solid particles in 12M HCl. 

The density was measured by weighing a known volume of OGaIn. A 1 $\times$ 1 $\times$ 1/4 inch mold was precision-cut from acrylic using a laser and then securely clamped to another acrylic piece. The weight of the empty mold was recorded. OGaIn was carefully applied into the mold using a wooden craft stick, ensuring thorough filling of corners and the absence of air gaps. Excess material was removed by drawing the craft stick across the mold's top. The filled mold was weighed again, and the OGaIn density was calculated by subtracting the empty mold weight from the filled mold weight and dividing by the known volume. 
We measured the density of three samples and reported the mean.

XRD testing was completed on a Rigaku SmartLab machine with a 2~mm window. SEM images were taken on a Hitachi CFE SU8230. EDS analysis was performed with a Bruker QUANTAX FlatQUAD (mapping at 5~kV and compositional analysis at 6~kV). Surface samples were prepared on 12.5~mm stubs on carbon tape. Cross-section samples were prepared on vertical sample holders on carbon tape, where a trace of OGaIn was frozen and then broken in half and mounted to the stub. Stretched cross sections were prepared by casting OGaIn on VHB, encapsulating with rubber cement, and then stretching the substrate while applying to the SEM stub. After mounting the sample, it was frozen and then broken to reveal the stretched cross section. 

The single-trace samples were created using the dimensions and electrode attachment techniques described in~\cite{Sanchez-Botero2022}, with VHB tape and rubber cement as the substrate and encapsulant, respectively. The trace widths of OGaIn were 250 \unit{\micro \meter}, with gauge length 25 mm (Fig.~S3
B,C). 0~$\Omega$ resistor samples were made using the same patterns and processes, but with a 0.25 mm gap in the middle of the trace, over which an 0402 0~$\Omega$ surface mount resistor was placed (Fig.~S3
E). The single trace samples were characterized using the custom setup previously described in~\cite{Sanchez-Botero2022}, and illustrated in Fig.~S3
D. An Arduino-driven lead-screw actuator controlled the strain and strain rate, as resistance was recorded using a 4-point probe multimeter (BK Precision).

\subsection*{Substrate viability characterization}
\label{methods: sub via}
ASTM D6195-22 standard tack testing was carried out using an Instron 3345 tensile tester (maximum capture rate 2~\unit{\milli s}, Fig.~S9
A). 
Five samples of each material were created as follows: VHB tape was adhered to the polyethylene terephthalate (PET) backing, ensuring no air bubbles were present. The DS10 (DragonSkin 10A, SmoothOn Inc.) samples were made by casting DS10 over the PET backing with a 0.5~mm draw bar. That same technique was used to create the samples of Slacker 1, 2 and 3. Slacker 1 used one part DS10 Part B, one part Slacker (SmoothOn, Inc.), and one part DS10 Part A (denoted, 100:100:100). Slacker 1.5 used the ratio 100:150:100, and Slacker 2 used 100:200:100.

The same setup, sample creation, and procedure were used for the OGaIn adhesion as for the previously described tack testing, though instead of coming into contact with steel, the samples came into contact with molded OGaIn (Fig.~S9
B-E). A 25 $\times$25~mm, 1/16~in deep mold was created to fit onto the steel bar used for previous experiments (Fig.~S9
B). This mold was filled with OGaIn before each new sample, and a bar was drawn across the top surface before removal of the mold (Fig.~S9
C). 

To get an application-based analysis of the effects of OGaIn-substrate adhesion on the circuit manufacturing process, defect rate tests were carried out. Five traces (250~\unit{\micro \meter} wide and 45 mm long) of OGaIn were screen printed onto each of the substrates (VHB tape, DS10, Slacker 1, 2, 3) using a sticker paper mask. The conductivity of each trace was measured, and a binary rating of conductive or nonconductive was recorded for each trace (Table~S1). 

\subsection*{Stretchable Arduino Pro Mini manufacturing}
\label{methods: circ mfg proc}
The circuit pattern for the Arduino Pro Mini was downloaded from Arduino.com in the industry-standard Eagle PCB-design file format. The relevant circuit layers---bottom, top, VIAs, and header pin holes---were exported to CorelDraw for post-processing  and to prepare the files for laser cutting (Fig.~3
A). In CorelDraw, ground lines were added to replace the commercial PCB's ``ground pour.'' VIAs and header pin hole sizes were reduced to 0.4 mm in diameter (Fig.~3
B). The reduction in VIA size from the open source design to the soft design left sufficient clearance between the VIAs and traces, as well as the header pins and traces.
The implemented circuits also featured a minimum trace width of 0.2~mm, a minimum clearance between VIAs and traces of 0.29~mm, and a minimum distance between traces of 0.17~mm.

Here we detail the manufacturing procedure illustrated in Fig.~3 
and recorded in Movie S7. 
Though Fig.~3 
presents a general procedure, this text will use VHB tape (3M Inc., VHB 4905) as the circuit substrate, and rubber cement (Elmer's, Inc.) as the encapsulant. If a DS10/Slacker circuit is desired, replace the VHB tape with a cured 0.5 mm casting of Slacker 1, 1.5, or 2, and replace rubber cement with uncured Slacker 1, 1.5, or 2. To compare each material's stiffness, we measured the 100\% modulus of all materials used in this manufacturing process following ASTM D412-16, with Dogbone type C (Fig.~S10
). Briefly, the modulus of SilPoxy $E_{sp}$ was $\sim$4$\times$ that of DS10 ($E_{DS10}$), which was similar to VHB and rubber cement, followed by Slacker 1, Slacker 1.5, and Slacker 2 being $<0.2 \times E_{DS10}$.

First, sticker paper from the back of double-sided tape (Amazon, Le Papillon Jewelry) was adhered to both sides of the VHB tape, and bubbles were removed using a flat edge of a piece of acrylic (Fig.~3
C). Then a CO$_2$ laser (ULS VLS2.30DT) was used to cut the board outline and the VIAs/header pin holes, using settings of 100\% power and 20\% speed, for two repetitions  (Fig.~3
D). Next, the top and bottom traces were cut into just the top and bottom pieces of circuit paper, respectively, with a UV laser (LPKF ProtoLaser U4, LPKF, Inc.). The CO$_2$ laser, with a beam width of approximately 300 $\mu$m, efficiently cut through the substrate but had relatively low resolution. Conversely, the UV laser, with a beam width of 15 $\mu$m, could cut precise, high-resolution mask traces but was unable to cut through the substrate. To ensure alignment, the board outline was cut into a piece of paper, and the circuit was then placed into the cutout in the paper top side up, and taped in place. The top traces were then engraved into the mask (Fig.~3
E). Flipping the circuit over its neutral axis and aligning in the paper cutout, the bottom traces were then engraved (Fig.~3
F). 

The pieces of sticker paper where the conductive material should go were removed with tweezers from the bottom side. The VIAs were filled with OGaIn using a sharp multimeter probe (FLUKE TP88). Then, OGaIn was painted into the mask using a paintbrush---brushing perpendicular to traces and in circles over them (Fig.~3
G). Finally, the edge of a piece of acrylic was used to scrape excess material off the top, and the mask was removed (Fig.~3
H). Each trace was tested for conductivity using a 2-point probe multi-meter (FLUKE-83-V) with the probes listed previously. To encapsulate, rubber cement was loaded into a syringe and extruded out of a 17~Ga needle over all exposed OGaIn (Fig.~3
I). 

Once the bottom side was cured, the same process was repeated until top mask removal, (Fig.~3
J), and components were placed exactly as they were on the rigid Pro Mini, using tweezers. Wires were then placed with tweezers and adhered with Sil-Poxy (SmoothOn Inc.) (Fig.~3
K). Next, a program was uploaded to the device to test for functionality. Finally, the top was encapsulated, and the Sil-Poxy was added over the microprocessor and oscillator to create a strain gradient. Once cured, the circuit was tested again. Notes on improving and scaling this process can be found in the Supplementary Methods.

The same process was used to make the other example circuits (Sparkfun Sound Detector, Arduino Lilypad, and Sparkfun RGB and Gesture Sensor), using their open source designs. An extremely fine layer of Sil-Poxy was added on the bottom layer under the RGB sensor IC, and no encapsulant or Sil-Poxy was added directly over the lens.

\subsection*{Circuit characterization}
\label{methods: circ char}
The five Arduino Pro Mini samples used for pull-to-failure tests were manufactured with a border on the edges such that fabric could be adhered, limiting the strain in those regions and ensuring that only the circuit board was straining. To set up, these fabric regions were gripped (Instron 2713-007) by the materials testing system (Instron 3345), and the wires for serial connection were plugged into an FTDI (Future Technology Devices International Limited) programming board (Sparkfun DEV-09716), which was connected to a laptop. A code that flashed four off-board LEDs and printed timestamps to the serial monitor was uploaded to the soft Pro Mini. The laptop and Arduino were recorded by an external camera setup. The Instron captured force and displacement data while it pulled upward at a rate of 15 mm/min until serial disconnect occurred. The initial length and length at failure of the circuit were calculated from the camera footage (with one initial frame, and one frame just before serial disconnect), and from these, engineering strains were calculated.
This same process was used to test the failure of the three additional example circuits, one sample each, though they were gripped with acrylic adhered to the circuit as the strain-limiter. The Arduino Lilypad was strained until the LED stopped blinking at the rate specified in the script. The Sparkfun Sound Detector was strained with loud ambient noise to detect failure time. The Sparkfun RGB and Gesture Sensor sent color output through serial until an error reading serial occurred. 
For all measurements in this paper, we chose to calculate the strain clamp-to-clamp (field standard~\cite{ozutemiz_egainmetal_2018,ozutemiz_scalable_2022,Xie2023,Huang_3D_circuit_2018,Lee-omnidirectional-2023}).

Cyclic testing of the soft Arduino Pro Minis used the same cyclic testing device as for the single-trace samples. The same additional grip areas were used, but instead of attaching fabric, laser cut pieces of 1/8" acrylic were pressed onto either side of the excess VHB tape such that the four through holes fit onto the alignment pins of the cyclic tester (Fig.~S11
). The cyclic tester strained to 100\% strain for 1000 cycles, while a time lapse camera monitored the laptop screen, which concurrently displayed the serial output from the soft circuit, and the cycle number from the testing device. The number of cycles the circuit survived was recorded as the cycle before an error was seen in the serial output of the stretchable Arduino. 

After pull-to-failure and cyclic testing until serial disconnect, the samples were evaluated using the 2-point probe FLUKE multimeter to determine where the failure occurred (Fig.~S7
C).

To determine the effect the circuitry had on the circuit substrate's force-strain relationship, the same strain-limiting fabric pieces were attached to five stretchable Pro Mini circuits and five pieces of plain VHB tape. The materials testing system gripped these sections and strained each sample to 100\% strain for 200 cycles, at 15 mm/min. Data from cycle 1, 10 (after Mullin's effect~\cite{Mullins1969}, and 200, straining and relaxing, were isolated for each sample. The means and standard deviations were then calculated for the five samples, from cycles 1, 10, and 200, straining or relaxing.

\subsection*{Fabrication and design of quadruped}
\label{methods: quad design}
The soft quadruped robot was inspired by Shepherd, et al. 2011~\cite{shepherd_multigait_2011}. The top portion of the robot was DS10 cast in a 3D printed mold. Once cured, this top portion of the robot was placed onto fabric impregnated with uncured DS10, and the two parts cured together, with the bottom fabric acting as a strain limiting layer. The bladders for the four legs and the body each had an exit to open air that, instead of being on the top, were out the sides. In these exits, silicone tubes (McMaster 5236K203) were inserted with tweezers, and secured with Sil-Poxy. 

To distribute the circuit design, the outline of the soft quadruped robot designed above was imported into AutoDesk Eagle as a board outline. The routing from the Arduino Pro Mini was then erased, the components manually moved to the desired locations on the board, and the auto-route feature in Eagle was used to create new routes, including ground lines. 

The DS10 quadruped circuit was fabricated using the same manufacturing procedure, but Slacker 1.5 was used as the substrate and encapsulant (Fig.~S12
A). For both the Slacker and VHB tape circuits, the I$^2$C communication wires were added after the top was already encapsulated to ensure that no traces were bridged during wire placement. To do this, tweezers pinched the encapsulant above the header pin hole, and sewing scissors were used to snip a hole in the encapsulant. The wires were then attached the same way as described previously, and a dab of extra encapsulant was put over the wire lead. 

The Slacker circuit was attached to the DS10 robot by coating a thin layer of DS10 over the top of the robot, and placing the circuit on top to cure (Fig.~S12
B). The VHB tape circuit adhered loosely to the robot, though the rubber cement did not adhere. At the tip of each leg, the circuit was mechanically bonded to the robot using DS10 (Fig.~S12
C,D).

The quadruped setup (Movies S3 and S4) was as follows: the soft Arduino was connected to the laptop for the power supply, and to the I$^2$C breakout board. The breakout board was connected to the pressure regulators. Compressed air fed into the pressure regulators, the outputs of which went to each leg and the body. The Arduino had a programmed gait that it ran by controlling the pressure regulators, using I$^2$C protocol.

\subsection*{Fabrication and design of voxel robots}
\label{methods: vox design}
The voxel robots were composed of 16 voxels, each dimensioned 15 $\times$ 15 $\times$ 60~cm, as fabricated and assembled in prior work~\cite{kriegman_automated_2019,kriegman_scalable_2020}. Sil-Poxy was used to secure tubes in each bladder, and apply friction-biased acrylic feet (Fig.~S13
) to ensure one dimensional (1D) motion over the muslin fabric surface. 

A stretchable LED array board was manufactured using the same method as the regular circuitry (Fig.~S8
), and consisted of three LEDs (red, yellow, and green) and three 220~$\Omega$ resistors. The grounds all connected to a point on the board that would be next to the ground pin on the Arduino, when placed side by side. The three lines from the LEDs also came to their ends directly across from a set of digital pins on the Arduino. This board design was used for this demo and the wearable demo.

The stretchable Arduino and LED board were placed next to each other on top of each voxel robot, with a $\sim$1~mm overlap between the sensing board and the Arduino, so that the two boards would not separate when strained. Using tweezers to pull up the encapsulant, and sewing scissors to cut it away, a digital pin, two grounds, and pin A3 were exposed. Using a multimeter probe, OGaIn was painted from the digital pin on the Arduino to the LED trace on the breakout board, and from ground to ground. These connections were then encapsulated. Next, the A3 pin was extended in the same manner, painting a path all the way to the top, front-facing voxel on the voxel bot, and covering that face (not encapsulating, to maintain conductivity). The same was done to connect the other ground to the bottom voxel. 

The soft Arduinos ran a code that spent half of the time listening for a 5~V signal, and half of the time sending a 5~V signal. This way, when the A3 and ground pins on each robot made contact with each other, they would light up the green LED. A valve connected to compressed air at 20 psi was repeatedly opened and closed to get the two robots to inch toward each other.

\subsection*{Fabrication and design of wearable system}
\label{methods: wearable}

A textile-based capacitive strain sensor, as developed in our previous work~\cite{Anjali_wearable}, was embedded into the elbow sleeve using the sleeve's fabric as a dielectric layer. The two leads from the sensor were soldered to a rigid capacitive sensor processing board (MPR121, Adafruit). From this board, 5V, GND, I$^2$C lines connected to the stretchable Arduino. Another LED breakout board was fabricated for this demonstration. The serial communication lines from the Pro Mini connected to the FTDI programming board, which connected through USB to a computer. 

Laying the sleeve flat on a table, the Arduino was placed directly over the top of the sensor, and pressed into place. The LED breakout board was placed next to it, so the ground and digital pins aligned, and the substrated overlapped by 1 mm. The header pin holes for the three digital pins, LEDs and ground were exposed on both the Arduino and the LED board, connected with more OGaIn, and then encapsulated with rubber cement. 

\subsection*{Statistical analysis}
In each experiment, we have noted the number of times an experiment was independently performed (N). Where N $>$ 2, shaded regions or error bars indicate one standard deviation. For fit lines, R$^2$ values are reported. Units are listed for all measurements.

\section*{Supplementary Materials and Methods}

\indent Supplementary Methods 

Figures S1-S13

Table S1

Movies S1-S7

\bibliographystyle{Science} 
\bibliography{references.bib}

\section*{}
\textbf{Acknowledgements:} We thank Lina Sanchez-Botero for providing material for preliminary experiments and engaging in discussions. \textbf{Funding:} This material is based upon work supported by the National Science Foundation under grant no. IIS-1954591. S.J.W. was supported by a NASA NSTGRO Fellowship (80NSSC22K1188). D.S.S. was supported by a NASA Space Technology Research Fellowship (grant no. 80NSSC17K0164). A.A. was supported by an NSF Graduate Research Fellowship, DGE-2139841. \textbf{Author contributions:} R.K.B. conceived the project. S.J.W. designed and executed all experiments, and led the writing of the paper. D.S.S. developed the initial manufacturing procedure. D.S.S. and R.K.B. contributed to the design of experiments and data interpretation. M.L. contributed to fabricating and executing the quadruped and VoxelBot demonstrations. A.A. fabricated the sensor and helped execute the wearable demonstration. All authors contributed to the preparation of the manuscript.  \textbf{Competing interests:} Authors D.S.S. and S.J.W are listed as inventors on US Continuation-in-part Application No. 18/148,500, which claims priority to US Patent Application No. 17/357,060, on which D.S.S. is also an inventor. \textbf{Data and Materials Availability:} All (other) data needed to evaluate the conclusions in this paper are present in the paper or Supplementary Materials. The data and code for this study have been deposited in Dryad database https://doi.org/10.5061/dryad.80gb5mkxf and additional resources, including design files, laser cutting files, a bill of materials, and step-by-step instructions for creating your own stretchable Arduino, are on GitHub at https://github.com/swoodman11/SoftArduino. 

\FloatBarrier

\noindent

\FloatBarrier

\noindent
Movie 1: \textbf{Stretchable Arduinos overview.} By utilizing a biphasic (solid-liquid) EGaIn-based conductor, we created stretchable versions of several popular open-sourced circuits, including the Arduino Pro Mini. In this video, we show how to manufacture and test the circuits, then embed them into soft robotic systems to aid in sensing and control.


\newpage
\clearpage
\begin{center}

\title{\LARGE \bf Stretchable Arduinos embedded in soft robots 
}

\author{Stephanie J. Woodman$^{1}$, Dylan S. Shah$^{1}$, Melanie Landesberg$^{1}$,\\ Anjali Agrawala$^{1}$, and Rebecca Kramer-Bottiglio$^{1}*$
\\
\normalsize{$^{1}$Department of Mechanical Engineering \& Materials Science, Yale University.}
\\
\normalsize{9 Hillhouse Ave, New Haven, CT 06511, USA.}

\normalsize{*To whom correspondence should be addressed; E-mail:  rebecca.kramer@yale.edu.}
}
\end{center}
\FloatBarrier
\renewcommand{\thepage}{S\arabic{page}} 
\renewcommand{\thesection}{S\arabic{section}}  
\renewcommand{\thetable}{S\arabic{table}}  
\renewcommand{\thefigure}{S\arabic{figure}}
\setcounter{figure}{0}
\setcounter{section}{0}
\setcounter{page}{1}
\newpage
\clearpage
\section*{Supplementary Materials}
\label{sec: Supporting Information}
\noindent
\indent Supplementary Methods 

Figures S1-S13

Table S1

Movies S1-S7

\newpage

\FloatBarrier
\subsection*{Material analysis}
\label{sup text: SEM}
X-ray diffraction analysis of OGaIn showed amorphous peaks---indicating low crystallinity (Fig.~S1
A)~\cite{Chen2023}. SEM images of the surface showed an expected surface oxide, with occasional indents (Fig.~S1
B(i)) or protrusions (Fig.~S1
B(ii) marking air inclusions at the imaged surface. We took SEMs of the unstretched and stretched (200\% strain) cross sections of OGaIn (Fig.~S1
C,D). The indents and protrusions in the stretched OGaIn retained their relatively spherical shapes, in contrast to the elliptical shapes observed when liquid inclusions are elongated in a silicone matrix~\cite{bartlett_high_2017}. We suspect that the hard gallium oxide shell~\cite{Battu2018} surrounding each air inclusion prevents their deformation upon stretching, which could be why we observe similar qualitative rheological properties to BGaIn. Energy dispersive X-ray spectroscopy (EDS) showed the expected high content of gallium, low but consistent quantities of indium in the bulk, and mixes of high and low concentrations of oxygen on the surface, the latter of which correlates to the crumpled surface texture in the SEM image. Compositional analysis on the EDS showed 73.8\% Ga, 23.6\% In, and 2.6\% O, with an electron interaction depth of 0.4~\unit{\mu m} over a radius of 0.2~\unit{\mu um}. This is a slightly higher oxygen concentration than we measured by dissolving oxide content and weighing (1.4 wt\%), which is likely due to the limited penetration depth of the EDS, resulting in over-sampling of the surface oxide.
\newpage

\FloatBarrier
\subsection*{Rheology}
\label{sup text: Rheology}
We conducted rheological characterizations of both OGaIn and EGaIn using a Discovery Hybrid Rheometer 2 from TA Instruments. The measurements were performed with upper and lower geometries consisting of 40~mm stainless steel plates, with gaps set at 1500~$\mu$m. Our experimental protocol closely follows that of Daalkhaijav et al.~\cite{daalkhaijav_rheological_2018}.
Samples (three of each OGaIn and EGaIn) were equilibrated to 25~$^{\circ}$C for three minutes using a Peltier plate, followed by pre-shearing at 1 $s^{-1}$ for 2~minutes. Amplitude tests were then conducted at 1 rad $s^{-1}$ from 0.05\% to 1000\% strain to determine yield stress and yield strain (Fig.~S2
A-D). Tests were terminated if the raw phase exceeded 150 degrees to prevent damage to the instrument.
A flow sweep from 0.01 to 10 rad $s^{-1}$ was performed to assess the shear rate-dependent viscosity of the samples (Fig.~S2
E). Our measured yield stress of EGaIn, approximately 48 Pa (averaged between the values from Fig.~S2
B and C), is consistent with the value reported by Daalkhaijav et al. (approximately 49.5 Pa when averaged using the same tests). OGaIn exhibited a substantially higher yield stress, measuring 1758 Pa (also averaged between the values from Fig.~S2
B and C), compared to EGaIn.
The yield strains for EGaIn and OGaIn were found to be 2.99\% and 2.46\%, respectively (Fig.~S2
B). The linear viscoelastic regions (LVR) were observed up to 6.24\% strain for EGaIn and 4.23\% strain for OGaIn (Fig.~S2
D).
Notably, the flow sweep revealed that OGaIn exhibits shear-thinning behavior, indicating its potential suitability for extrusion printing.
\newpage

\FloatBarrier
\subsection*{Contact Resistance}
\label{sup text: contact resistance}
To determine the contact resistance of OGaIn samples interfacing with ICs, we employed the transmission line method~\cite{reeves_obtaining_1982} (Fig.~S6
). 
We compared neat OGaIn traces (three samples) to OGaIn traces with a 0~$\Omega$ resistor bridging a gap in the trace (three samples). In both cases, we used a 10~cm-long trace and measured electrical resistance using a BK Precision 4-point multimeter, gradually decreasing the probe separation and, hence, sample length). A line was fitted to the data, with the y-intercept representing the contact resistance.

The average contact resistance of the three neat specimens was 0.0007~$\Omega$. The average contact resistance of the three 0~$\Omega$ resistor samples was 0.04~$\Omega$.
For context, consider the sample from Fig.~2
B,C with conductivity $\sigma$ = 2.11 $\times 10^6$~S/m, length L = 0.025~m, and cross-sectional area A = 0.00025~m $\times$ 0.0001~m (width 250~$\mu$m, height 100~$\mu$m). The calculated resistance is $R=L/(A \sigma)$ = 0.47~$\Omega$, which is close to the measured resistance of $0.5~\Omega$ and indicates that the contact resistance between OGaIn and other components is on the order of tens of m$\Omega$.
\newpage

\FloatBarrier
\subsection*{Polymer Stiffness}
\label{sup text: polymer stiffness}
To assess the stiffness and stretchability of the polymers used in this study, we conducted uniaxial tensile tests on all seven materials (DS10, Sil-Poxy, rubber cement (RC), VHB, Slacker~1, Slacker~1.5, and Slacker~2), as shown in Figure~S10
. For each material, three samples of ASTM D412-16 Type C dogbones were prepared. The target thickness was 3~mm, but the thickness of rubber cement varied between 2-3 mm due to its high shrinkage ratio during solvent evaporation. The VHB used in the experiment was purchased at a thickness of 0.5 mm.
The samples underwent strain using a mechanical testing machine (Instron 3345) according to ASTM D412-16, generating the curves depicted in Figure S10
A,B, which were used to calculate the 100\% modulus values (Figure~S10
C). As expected, Sil-Poxy, our strain-limiting material, exhibited substantially higher average stiffness (430 kPa) compared to our substrates and encapsulants. At the other end of the spectrum, the Slacker variations showed low stiffness ($<$20 kPa), compared to DS10 and VHB. Rubber cement showed early failure at approximately 350\% strain at the characterized thicknesses. However, in the actual circuits, the rubber cement layer is less than 92~$\mu$m (averaged across three locations), as determined using a microscope height tool (Zeiss Smartzoom 5).
In addition to the standard 3~mm samples, we also tested 92~$\mu$m thick samples in the Instron, which did not fail in the strain range tested (up to 400\% strain).
\newpage

\FloatBarrier
\subsection*{Further limitations and future work}
\label{sup text: lims} Although the stretchable circuits are robust enough to survive more than 12 hours of cyclic testing, there is still room for improvement. Interfacing around the 32-pin microprocessor still relies on the incorporation of higher-modulus silicone around the IC, to minimize the stress concentrations and trace bridging around the processor. We found that the adhesion between the substrate and OGaIn correlated with the tack of the substrate, suggesting that candidate materials can be identified utilizing standardized tack tests such as ASTM D6195-22. In future studies, we would like to extend this characterization to include surface modifiers, such as silanes and stretchable epoxies, and test the interfacial shear strength of the IC-substrate interface. This could improve the stretchability of circuits with IC components of every package type. Other possible solutions include the incorporation of materials with stiffness gradients to alleviate stress concentrations.
With these modifications, the circuit's robustness could further increase, especially under cyclic strain conditions. Future studies will characterize the circuits under 2D strain conditions, more closely mimicking the conditions they experience when integrated into soft systems. Additionally, though we used QFL (quad flat), SSOP (shrink small-outline), and DFN (dual flat no-lead) packages in this work, we aim to make our method compatible with other types like ball grid arrays (BGAs), quad flat no-lead (QFNs), and land grid arrays (LGAs) for future Bluetooth chips and other dense ICs, as this will be another step towards untethering soft systems with embedded soft electronics.
\newpage

\FloatBarrier
\begin{figure*} 
    \centering
    \includegraphics[width=6 in]{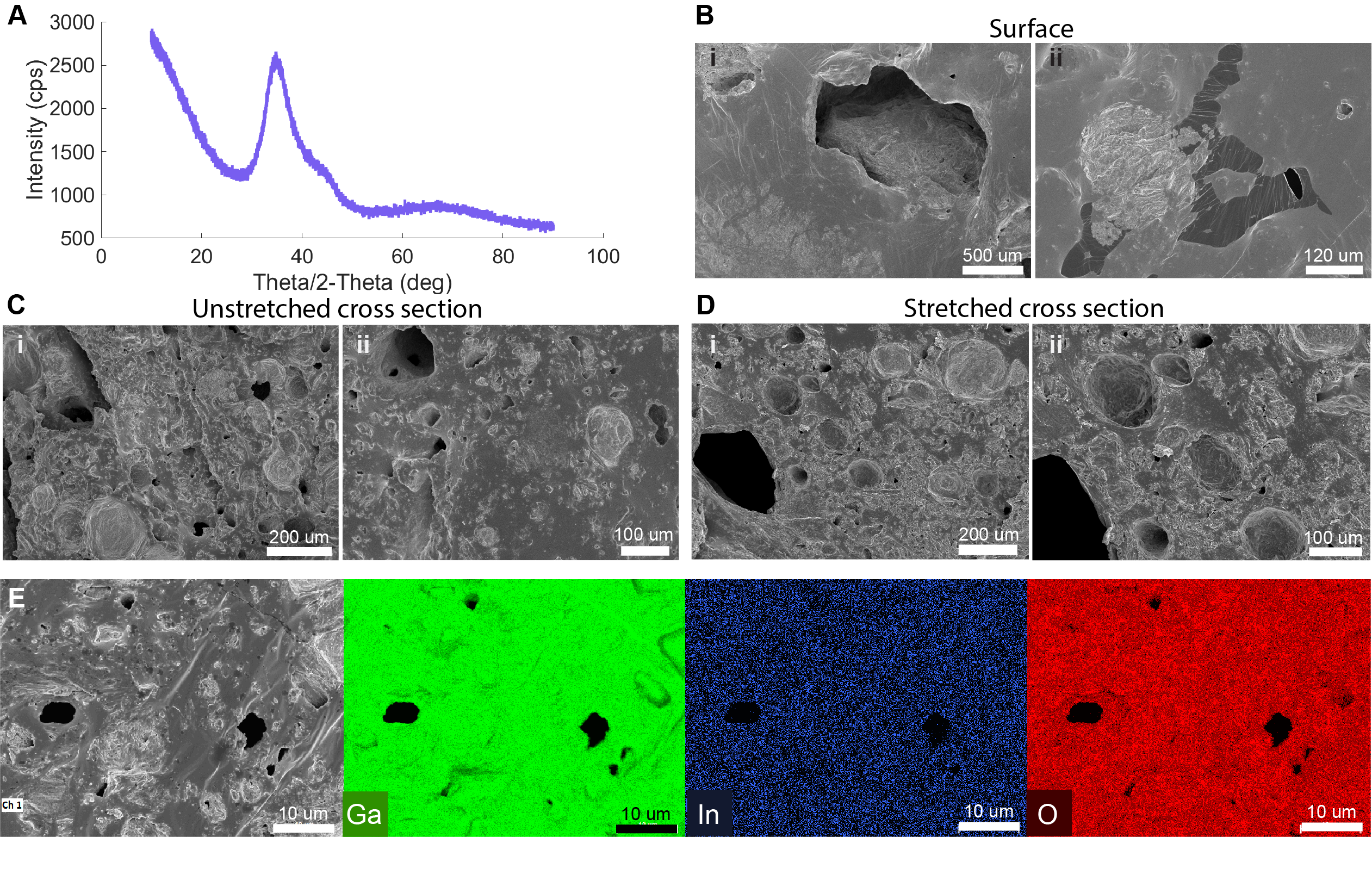}
    \caption{\textbf{XRD, SEM, and SEM/EDS analysis of OGaIn.}
    (\textbf{A}) XRD specturm of OGaIn. The lack of sharp peaks indicates a relative absence of crystalline gallium oxide, in favor of amorphous oxides~\cite{Chen2023,Liu_Bgain_2021}.
    (\textbf{B}) SEM images of the surface of OGaIn. Two types of occlusions are visible---a hollow oxide shell (i) and a crumpled gallium-oxide mass (ii).
    (\textbf{C}) SEM images showing the unstretched cross-section of OGaIn. Shows dense (i) and more sparse (ii) inclusions. 
    (\textbf{D}) SEM images of the stretched cross-section of OGaIn. Zoomed out (i) and zoomed in (ii), to match \textbf{(c)}. Note that the hollow shell and crumpled gallium-oxide inclusions both appear approximately the same shape as when unstretched.
    (\textbf{E}) EDS analysis of OGaIn cross-section. Gallium dominates the surface, along with visible surface indium and oxygen which is likely in the form of gallium oxide.
    }
    \label{sup fig: SEM}
\end{figure*}
\newpage

\FloatBarrier
\begin{figure*} 
    \centering
    \includegraphics[width=6 in]{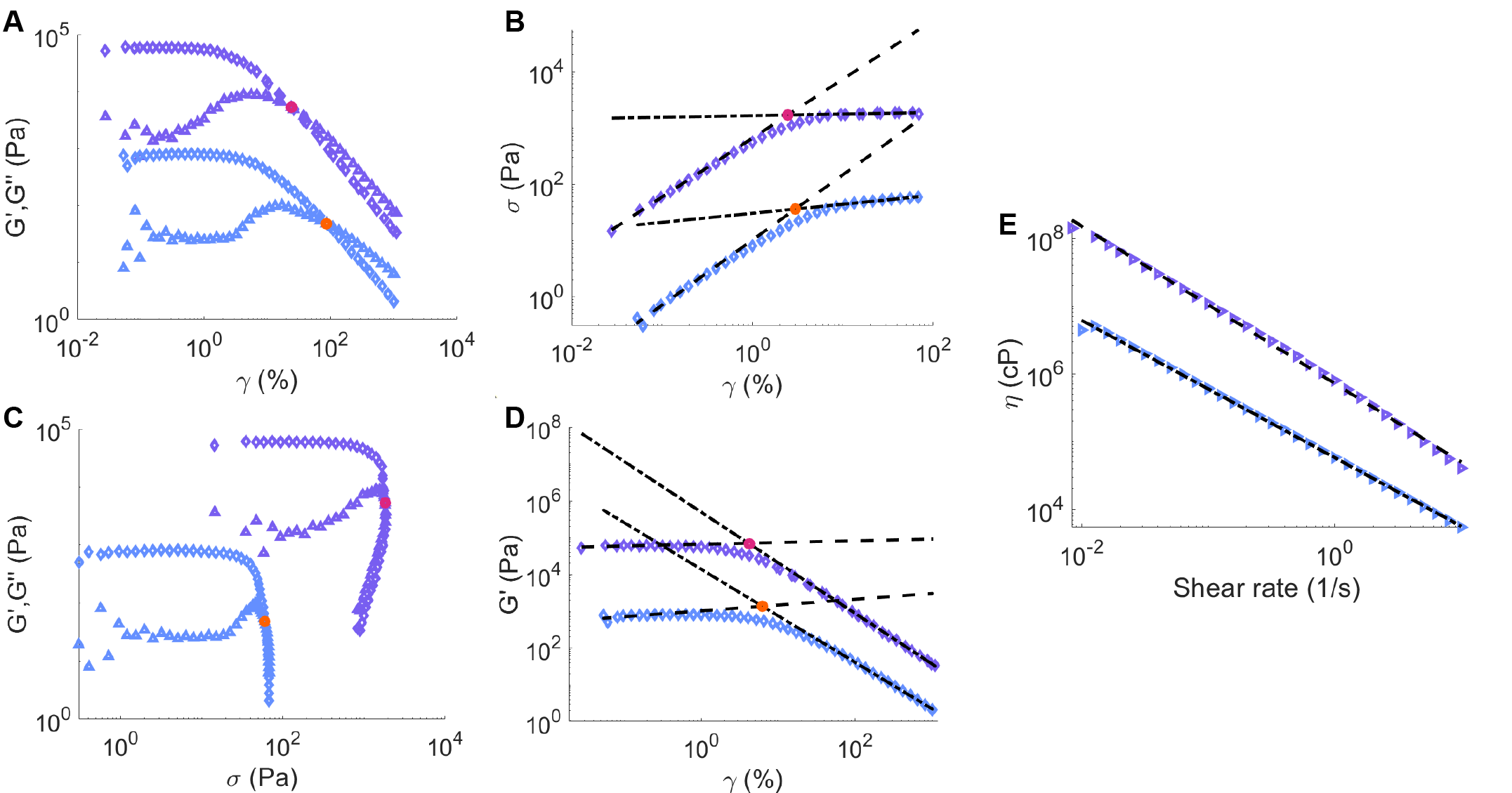}
    \caption{\textbf{Rheology of OGaIn.} Purple and blue markers indicate OGaIn data and EGaIn data, respectively. 
    (\textbf{A}) Storage modulus (G', diamonds) and loss modulus (G'', triangles) vs. strain for OGaIn and EGaIn. The resulting flow strains are 24\% for OGaIn and 87\% for EGaIn.
    (\textbf{B}) Stress vs. strain for OGaIn and EGaIn, giving yield strains of 2.46\% for OGaIn and 2.99\% for EGaIn and yield stresses of 1681~Pa for OGaIn and 36.1~Pa for EGaIn. The linear fits used to calculate yield strains were created using the first and last 8 data points for each curve, and are shown in black dashed lines. R$^2$ for the start and end of the purple line are 0.999 and 0.971, respectively. R$^2$ for the start and end of the blue line are 0.999 and 0.999, respectively. 
    (\textbf{C}) Storage modulus and loss modulus vs. stress, giving yield stresses of 1835~Pa for OGaIn and 60.13~Pa for EGaIn. Averaging these values with those from (B), the average yield stresses are 1758~Pa for OGaIn and 48.12~Pa for EGaIn.
    (\textbf{D}) Storage modulus vs. strain for OGaIn and EGaIn. Pink and orange dots indicate the location of the crossover point for OGaIn and EGaIn, respectively, which represents the end of the linear viscoelastic region (LVR). For OGaIn, this is 4.23\%, and for EGaIn this is 6.24\%. The linear fits used to calculate LVR were created using the first and last 8 data points for each curve, and are shown in black dashed lines. R$^2$ for the start and end of the purple line are 0.989 and 1.000, respectively. R$^2$ for the start and end of the blue line are 0.989 and 1.000, respectively.
    (\textbf{E}) Viscosity vs. shear rate, indicating that both OGaIn and EGaIn are shear thinning and could be suitable for extrusion printing. The linear fits approximating the data are shown in black dashed lines, with R$^2$ of the fit for the purple line being 0.998, and R$^2$ of the fit for the blue line being 0.999. 
    }
    \label{sup fig: rheology}
\end{figure*}
\newpage
\begin{figure*} 
    \centering
    \includegraphics[width=6 in]{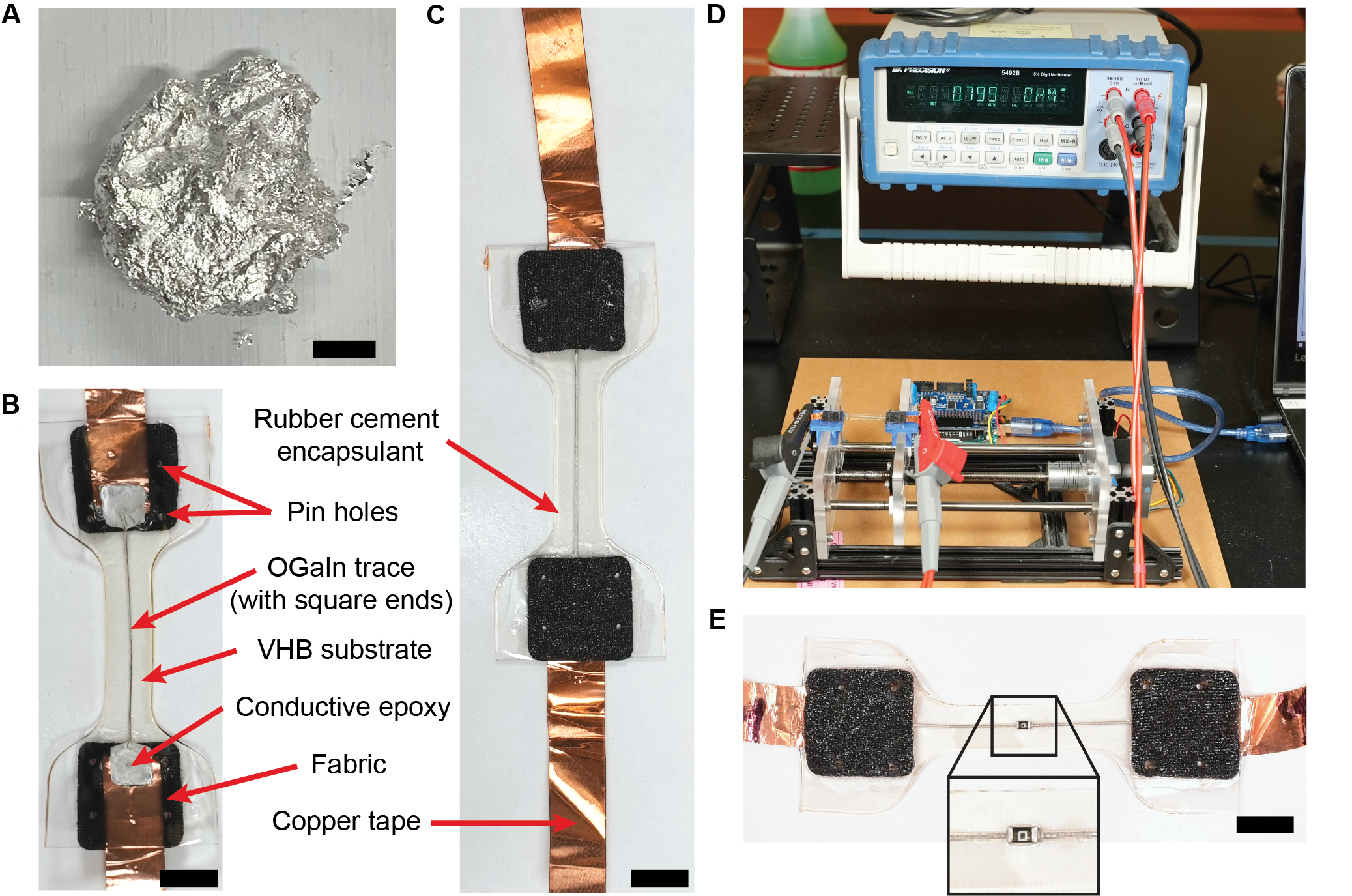}
    \caption{\textbf{OGaIn paste, images of single-trace samples, and cyclic tester.} 
    (\textbf{A}) Image of OGaIn. Scale bar 8 mm. The presence of gallium oxide allows the specimen to retain its shape, with a paste-like rheology.
    (\textbf{B}) Single-trace sample from the bottom. Scale bar 6.5 mm. 
    (\textbf{C}) Single-trace sample from the top. Scale bar 6.5 mm.
    (\textbf{D}) Single-trace sample being strained on the cyclic testing setup described in Sanchez-Botero \textit{et al.} 2022~\cite{Sanchez-Botero2022}.
    (\textbf{E}) Image of the single-trace sample with 0~$\Omega$ resistor. Scale bar 6.5 mm. The only difference between this sample and the one shown in (\textbf{B-C}) is the presence of the 0~$\Omega$ resistor bridging an intentionally-included gap in the OGaIn trace, which allows for the characterization of the interfacial resistance of IC components during stretch.
    }
    \label{sup fig: OGaIn sts}
\end{figure*}

\begin{figure*} 
    \centering
    \includegraphics[scale=0.75]{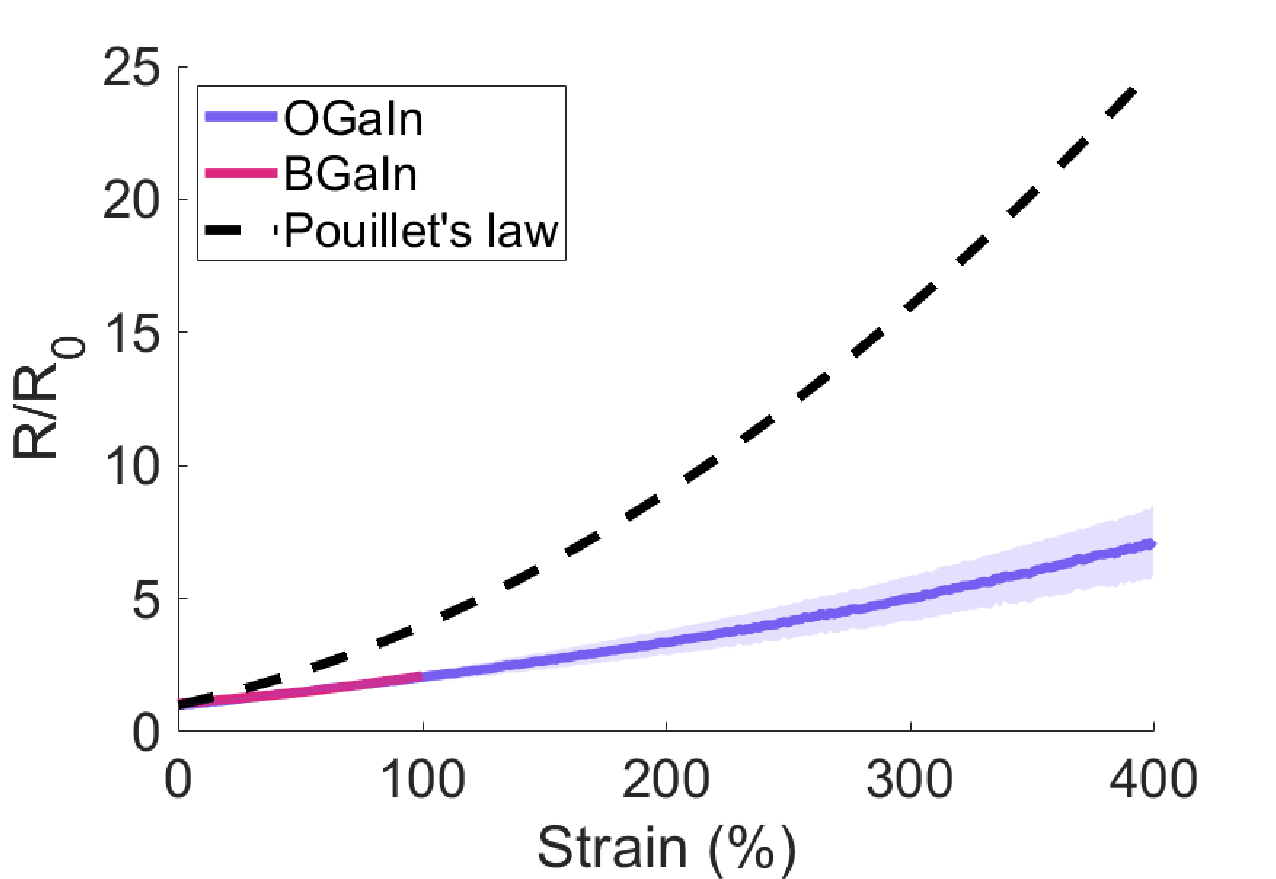}
    \caption{\textbf{Direct comparison between OGaIn and BGaIn.} Normalized resistance vs. strain for five single-trace samples each of BGaIn~\cite{Sanchez-Botero2022} and OGaIn, patterned on DS10 with a 0.8~mm trace width (differing from Fig.~2
    A, where the OGaIn trace width was 0.25~mm, to match the trace width of the Arduino Pro Mini). The curves overlap in the region where both could be compared. The shaded area is one standard deviation, and the purple line is the mean, of the 5~samples. 
    }
    \label{sup fig: direct comp}
\end{figure*}

\begin{figure*} 
    \centering
    \includegraphics[scale=0.75]{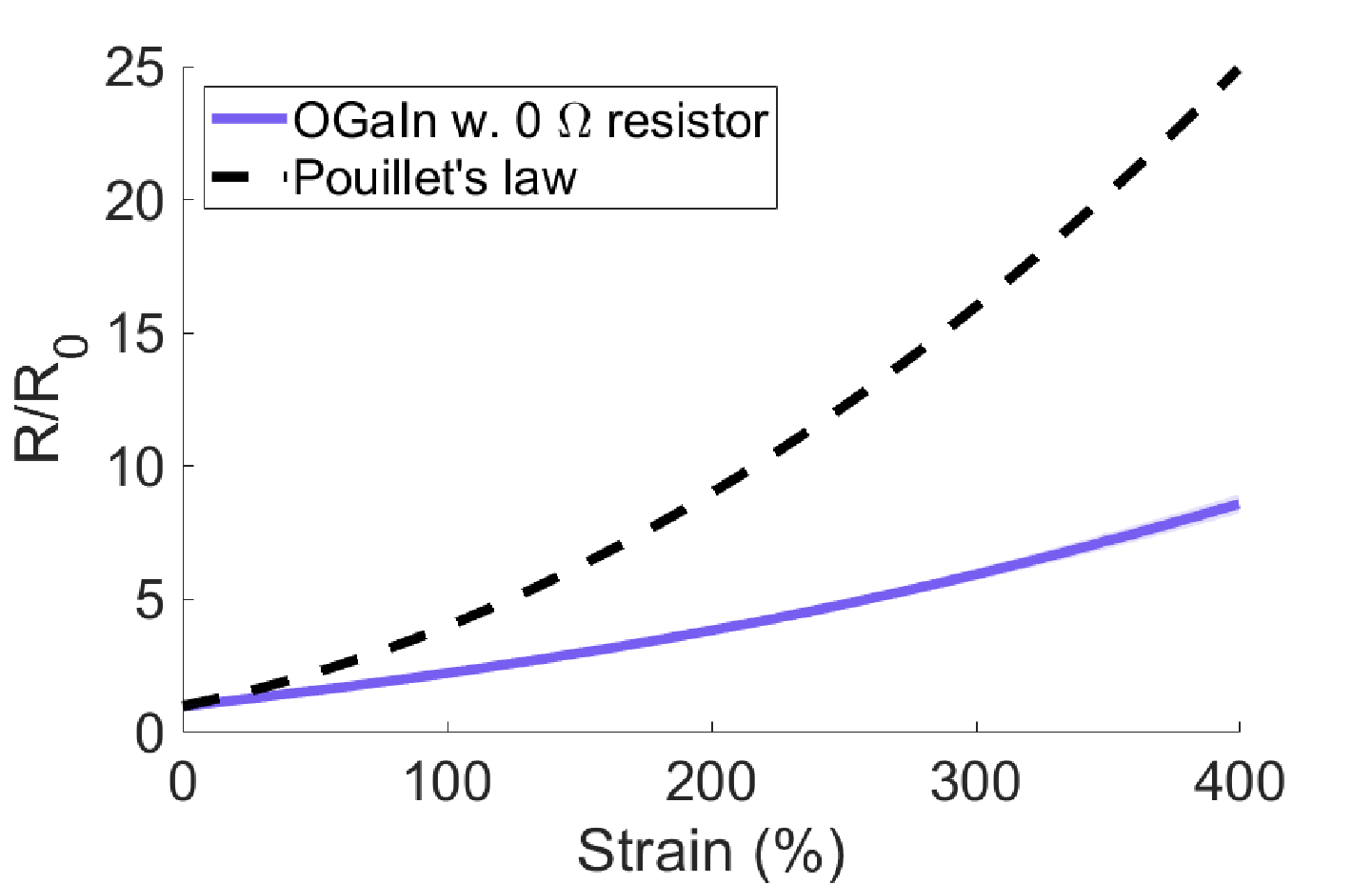}
    \caption{\textbf{Normalized resistance vs. strain for five single-trace samples with 0~$\Omega$ resistors.} The shaded area is one standard deviation, and the pink line is the mean, of the 5 samples. Pouillet's law is the dashed black line.
    }
    \label{sup fig: 400 0ohm}
\end{figure*}

\begin{figure*} 
\centering
    \includegraphics[scale=1.3]{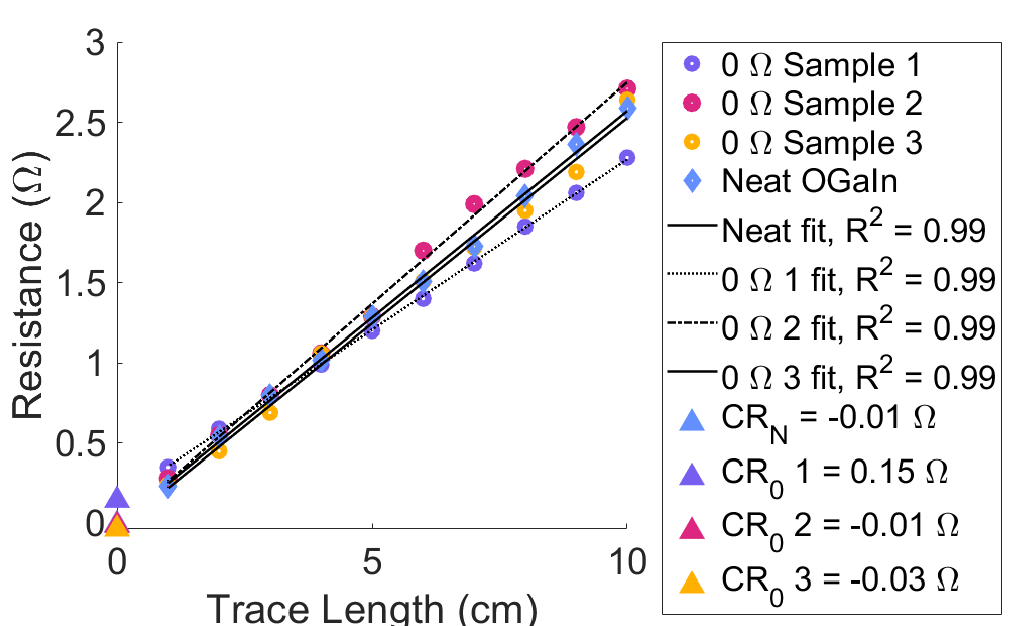}
    \caption{Contact resistance measurements for a representative neat OGaIn single-trace sample and three 0 $\Omega$ resistor samples. Linear fits for the data from each sample are provided along with their R$^2$ values. The intersection of these lines with the y-axis were used to get the contact resistance (CR).}
    \label{sup fig: transm line}
\end{figure*}
\newpage
\begin{figure*} 
    \centering
    \includegraphics{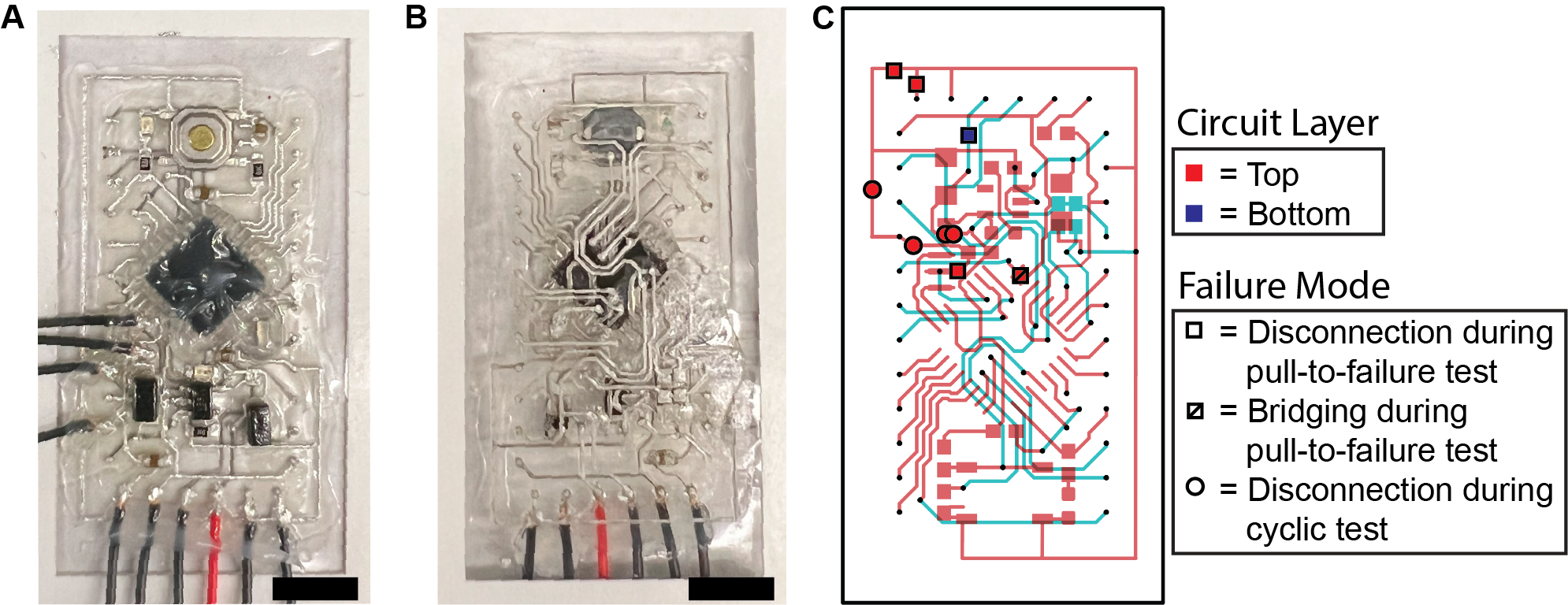}
    \caption{\textbf{Circuit images and pull-to-failure serial connection failure locations for the stretchable Arduino Pro Mini.} 
    (\textbf{A}) Top of completed VHB tape circuit. Scale bar 7 mm.
    (\textbf{B}) Bottom of completed VHB tape circuit. Scale bar 7 mm.
    (\textbf{C}) Locations where the circuits electrically failed after pull-to-failure testing and cycle-to-failure testing. Traces either failed due to a disconnection (trace is broken and creates an open circuit) or through bridging (two traces short circuit). Red-filled shapes indicate a failure on the top layer, as blue-filled shapes indicate a failure on the bottom layer. Empty squares indicate a failure by disconnection during the pull-to-failure test, squares with a diagonal line indicate a failure by bridging during pull-to-failure, and circles indicate a failure by disconnection during cyclic testing. 
    }
    \label{sup fig: circ images and SCF}
\end{figure*}

\begin{figure*} 
    \centering
    \includegraphics{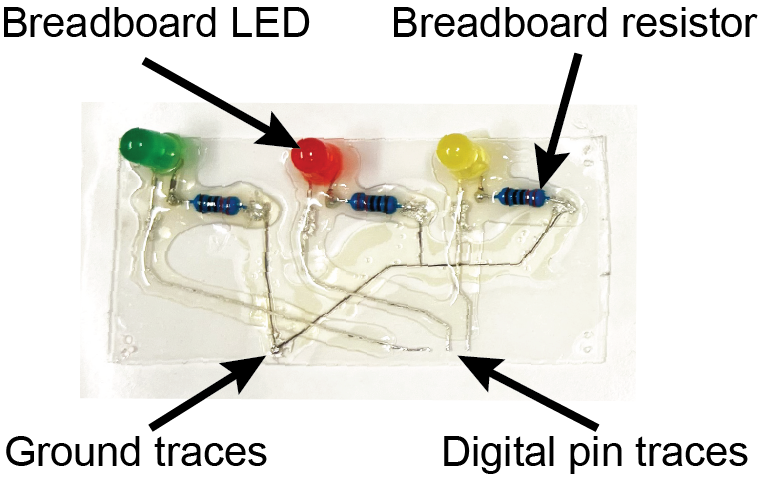}
    \caption{\textbf{Stretchable LED board for neighbor-neighbor contact sensing in VoxelBots, and co-located computation and sensing in a wearable system.} When the VoxelBots come in physical contact, the Green LED lights up every second. The same board design was used for the wearable example, where the green LED lit up at low strains, the yellow LED at medium strains, and the red LED at high strains.
    }
    \label{sup fig: LED bb}
\end{figure*}

\begin{figure*} 
    \centering
    \includegraphics[width=6 in]{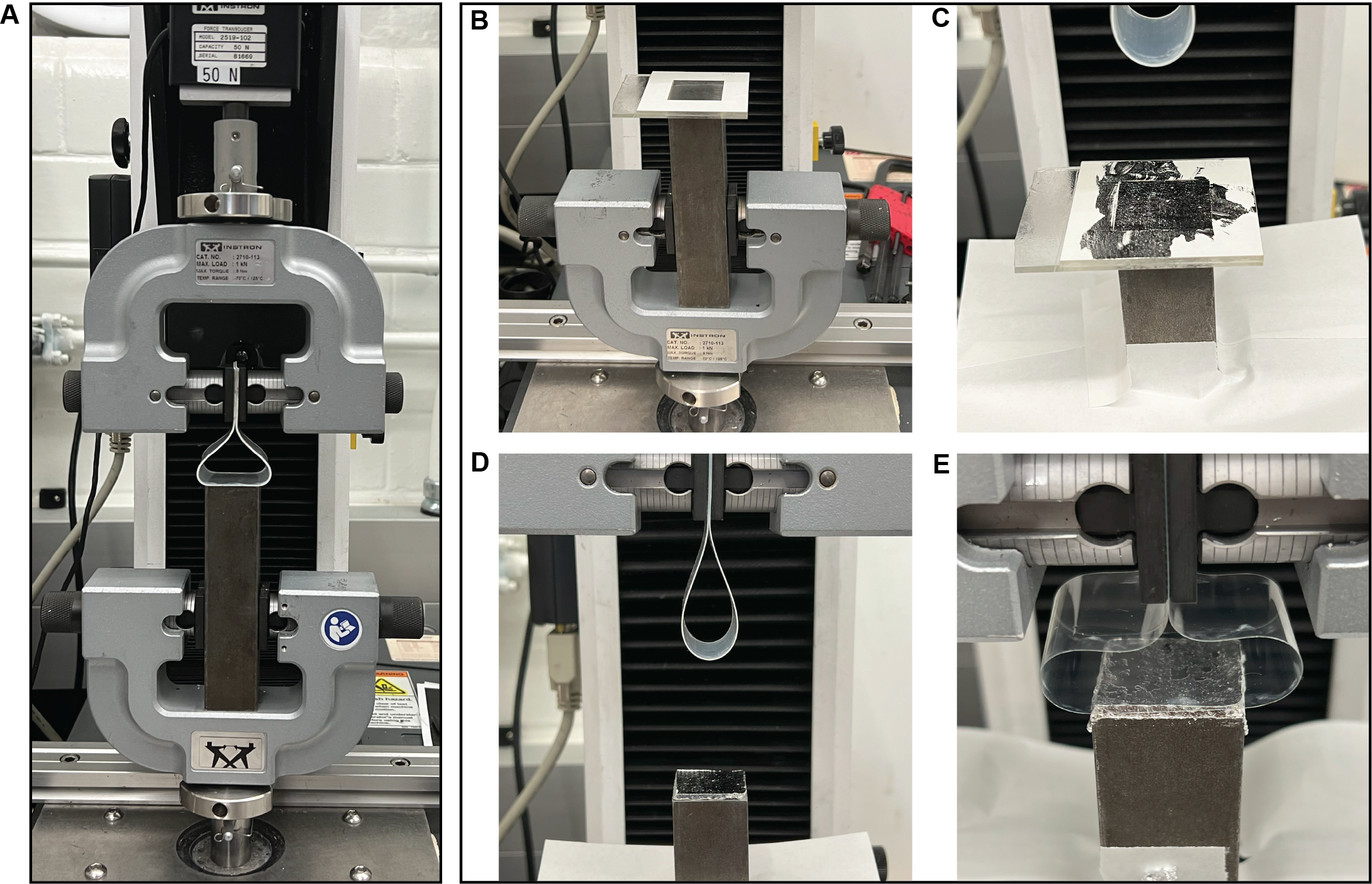}
    \caption{\textbf{Tack testing and OGaIn adhesion setup.} 
    (\textbf{A}) Image of ASTM D6195-22 tack testing setup mid-experiment, near when the loop makes full contact with the steel bar.
    (\textbf{B}) OGaIn mold seated on steel bar.
    (\textbf{C}) Filling the mold and scraping off excess OGaIn to make a flat surface.
    (\textbf{D}) After mold removal, with sample ready to make contact.
    (\textbf{E}) An example of a piece of PET making contact with the molded OGaIn. From here, the remaining steps are to lift the PET and record the force during the PET-OGaIn separation.
    }
    \label{sup fig: tack test and OGaIn}
\end{figure*}

\begin{figure*} 
    \centering
    \includegraphics[width=6 in]{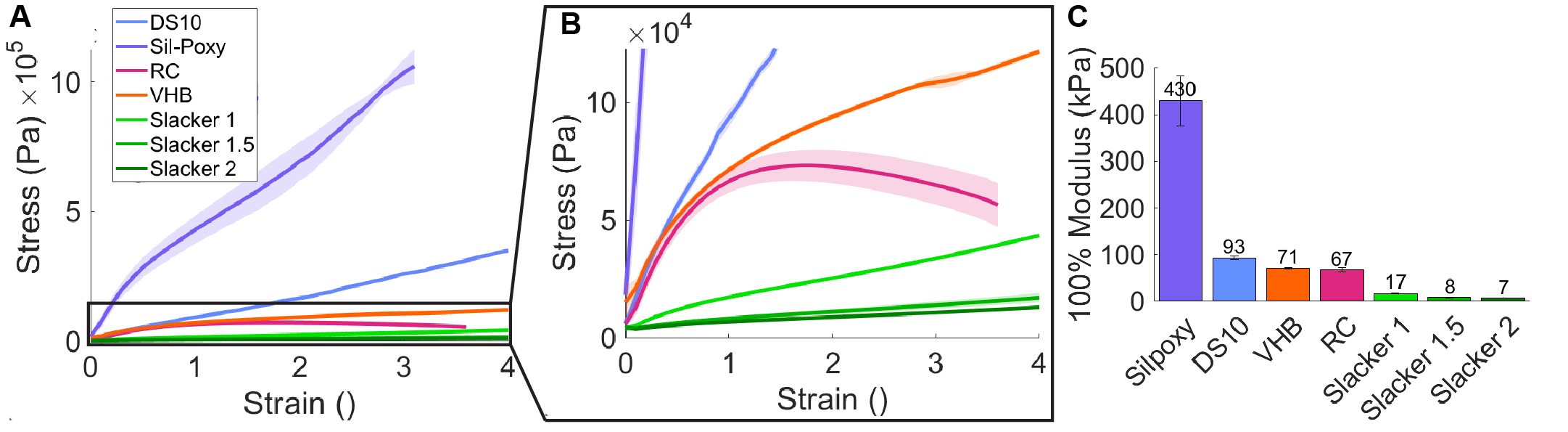}
    \caption{\textbf{Tensile testing the polymers used in circuit manufacture.} 
    (\textbf{A}) Stress vs. strain for all materials. Three samples were tested of each material. The shaded region represents one standard deviation. 
    (\textbf{B}) Stress vs. strain for all materials, zoomed in to show the trends of VHB, RC, Slacker 1, Slacker 1.5, and Slacker 2. The shaded region represents one standard deviation. 
    (\textbf{C}) Bar plot showing the 100\% modulus for each material, for comparison. Three samples were tested for each material, and the error bars represent one standard deviation.}
    \label{sup fig: 100 mod}
\end{figure*}
\newpage

\begin{figure*} 
    \centering
    \includegraphics{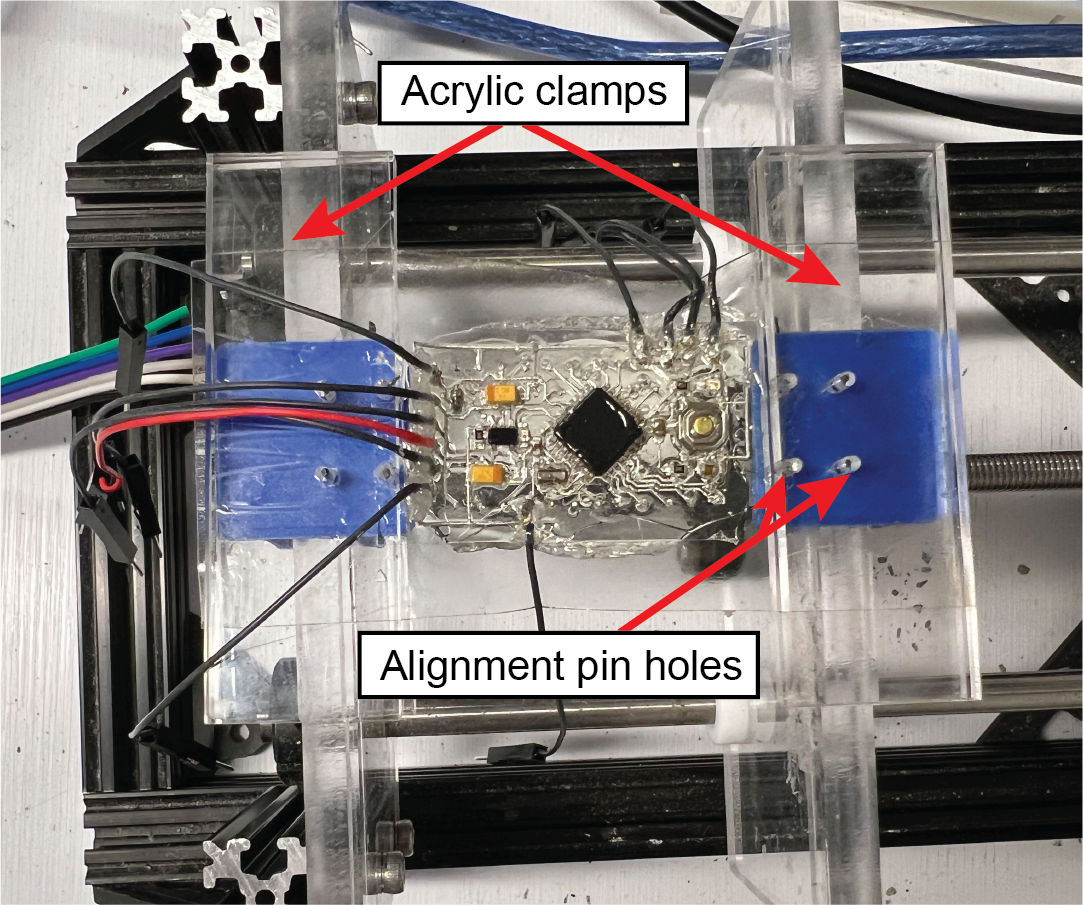}
    \caption{\textbf{Top view of the cyclic tester's specimen mounting mechanism.} The specimen's pull tabs are placed over alignment holes, and gripped using acrylic clamps. These two mechanisms help secure the specimen in place and isolate strain to the stretchable circuit area.
    }
    \label{sup fig: cyc test grips}
\end{figure*}

\begin{figure*} 
    \centering
    \includegraphics[width=6 in]{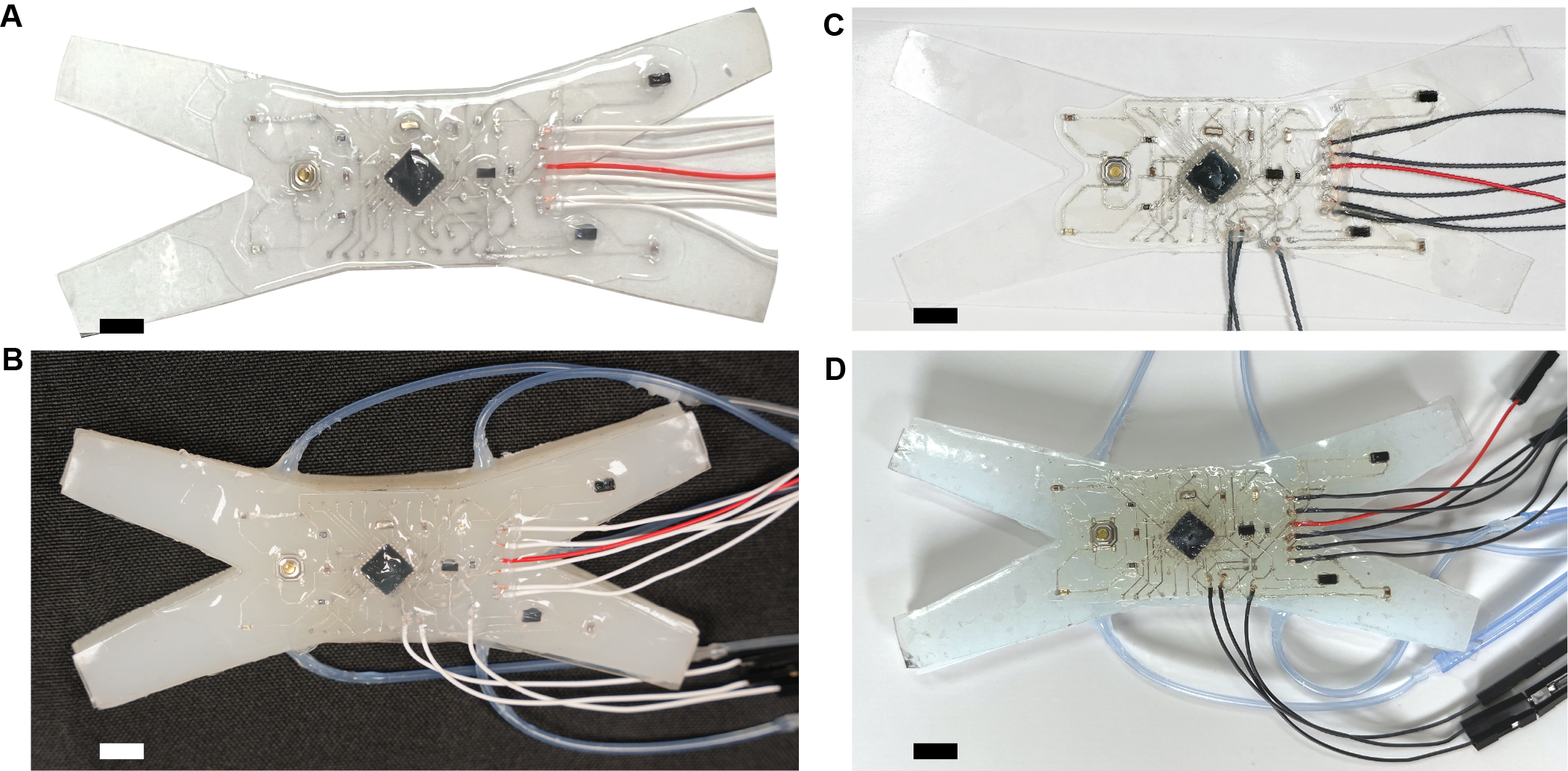}
    \caption{\textbf{Quadruped with integrated Stretchable Arduinos, made with DS10 and VHB substrates.}
    (\textbf{A}) ``Quadrupedal'' stretchable Arduino on a DS10 substrate, and (\textbf{B}) integrated into a DS10 quadrupedal robot.
    (\textbf{C}) ``Quadrupedal'' stretchable Arduino on a VHB substrate, and (\textbf{D}) integrated into a DS10 quadrupedal robot.
    In (\textbf{A-D}), scale bars are 7~mm.
    }
    \label{sup fig: quad circ}
\end{figure*}

\begin{figure*} 
    \centering
    \includegraphics{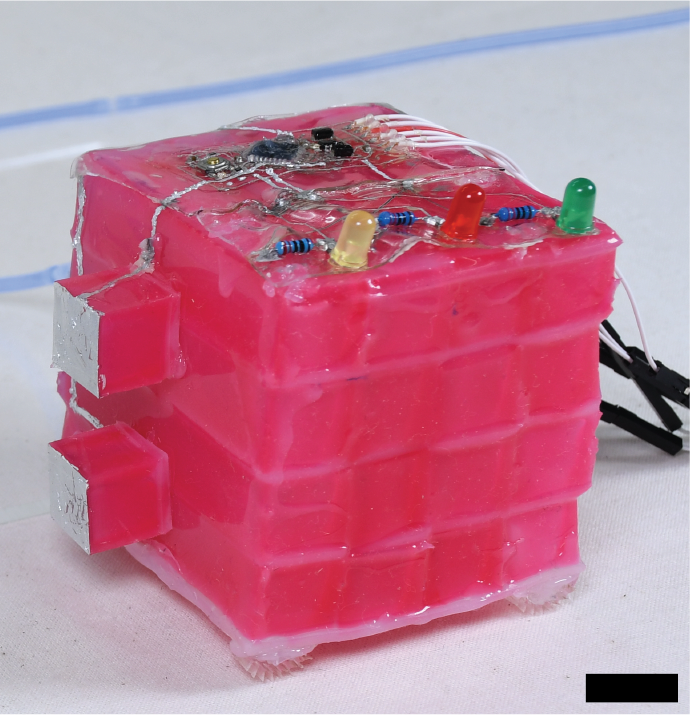}
    \caption{\textbf{VoxelBot with embedded circuitry and friction-biased feet.} Close-up of a VoxelBot with integrated LED indicator circuit and stretchable Arduino on top, and friction-biased feet on the bottom. Scale bar 15~mm.  
    }
    \label{sup fig: vox feet}
\end{figure*}

\FloatBarrier
\newpage
\begin{table}[h!]
\begin{center}
\begin{tabular}{||c c c c c c||} 
 \hline
 & DS10 & Slacker 1 & Slacker 1.5 & Slacker 2 & VHB tape \\ [0.5ex] 
 \hline\hline
 Trace failure rate & 1/5 & 1/5 & 0 & 0 & 0 \\ 
 \hline
\end{tabular}
\end{center}
\caption{\textbf{Trace failure rates for various substrates.}
    Five, 4.5~cm long, 250~\unit{\mu m} wide traces were applied using the same technique as for making a circuit to each of the five substrates. A binary ``conductive'' or ``non-conductive'' was recorded for each trace. This underscores the necessity for tack in the substrate to achieve a viable manufacturing process. 
    }
\label{sup tab: error rates}
\end{table}

\FloatBarrier
\noindent
Movie S1. \textbf{Pull-to-failure Instron testing.} This movie shows a stretchable Arduino Pro Mini during pull-to-failure testing, where the failure is a loss of serial connection to the computer. 
\newline
\newline
Movie S2. \textbf{Other circuits.} Here, we show three other open-source commercial circuits that have been translated to stretchable forms and their functionality. First is another microprocessor circuit, second is a sound detector, and third is an RGB and gesture sensor. 
\newline
\newline
Movie S3. \textbf{Spatially distributed stretchable Arduino controlling a soft robot’s gait (DS10).} A quadrupedal soft robot crawls across a fabric surface, being controlled by its surface-embedded stretchable Arduino Pro Mini (with a DS10 silicone substrate). The onboard stretchable Arduino maintains functionality while being stretched over 100\%, without restricting the robot's motion.
\newline
\newline
Movie S4. \textbf{Spatially distributed stretchable Arduino controlling a soft robot’s gait (VHB).} A quadrupedal soft robot crawls across a fabric surface, being controlled by its surface-embedded stretchable Arduino Pro Mini (with a VHB tape substrate). The onboard stretchable Arduino maintains functionality while being stretched over 100\%, without restricting the robot's motion.
\newline
\newline
Movie S5. \textbf{Neighbor-neighbor contact sensing in a soft, multi-agent system.} Two voxel robots crawl toward each other, and use their onboard stretchable Arduinos to sense and display contact.
\newline
\newline
Movie S6. \textbf{Wearable soft Arduino.} Here, a user wears a smart garment with an embedded Arduino Pro Mini and a fabric-based stretch sensor embedded in the area of highest strain. The stretchable Arduino Pro Mini processes the capacitive sensor data and displays bend ranges live using LEDs, giving visual feedback to the wearer.
\newline
\newline
Movie S7. \textbf{Detailed fabrication procedure.} This movie shows the manufacturing procedure presented in this work, to aid in reproducibility.


\end{document}